\documentclass[journal,twoside,web]{ieeecolor}
\usepackage{tmi}
\usepackage{cite}
\usepackage{amsmath,amssymb,amsfonts}
\usepackage{algorithmic}
\usepackage{graphicx}
\usepackage{textcomp}

\usepackage{multirow}
\usepackage{colortbl}
\usepackage{wasysym}
\usepackage{booktabs}
\usepackage{makecell}
\usepackage{color}
\usepackage{wrapfig}
\usepackage{array}
\usepackage{colortbl}
\usepackage{threeparttable}
\definecolor{myblue}{RGB}{194,228,248}
\usepackage{hyperref}

\DeclareUnicodeCharacter{2212}{-}

\usepackage{xcolor}

\def\BibTeX{{\rm B\kern-.05em{\sc i\kern-.025em b}\kern-.08em
    T\kern-.1667em\lower.7ex\hbox{E}\kern-.125emX}}
\begin{document}
\title{
HoloDx: Knowledge- and Data-Driven Multimodal Diagnosis of Alzheimer’s Disease
}

\author{Qiuhui Chen, Jintao Wang, Gang Wang, and Yi Hong
\thanks{This research work was supported by the National Natural Science Foundation of China (NSFC) 62203303, National Key R\&D Program of China 2024YFC2511100/2024YFC2511105 and Shanghai Municipal Science and Technology Major Project 2021SHZDZX0102.}
\thanks{Qiuhui Chen and Yi Hong are with the School of Computer Science, Shanghai Jiao Tong University, Shanghai 200240, China. Jintao Wang and Gang Wang are with the Department of Neurology, Renji Hospital Affiliated to Shanghai Jiao Tong University School of Medicine, Shanghai 200025, China.}
\thanks{Qiuhui Chen and Jintao Wang are co-first authors; Yi Hong (e-mail: yi.hong@sjtu.edu.cn) and Gang Wang (e-mail: wgneuron@hotmail.com) are co-corresponding authors.}
 }


\maketitle

\begin{abstract}

Accurate diagnosis of Alzheimer's disease (AD) requires effectively integrating multimodal data and clinical expertise. However, existing methods often struggle to fully utilize multimodal information and lack structured mechanisms to incorporate dynamic domain knowledge. To address these limitations, we propose HoloDx, a knowledge- and data-driven framework that enhances AD diagnosis by aligning domain knowledge with multimodal clinical data. HoloDx incorporates a knowledge injection module with a knowledge-aware gated cross-attention, allowing the model to dynamically integrate domain-specific insights from both large language models (LLMs) and clinical expertise. 
A memory injection module with prototypical memory attention further enables consistency preservation across decision trajectories. Through synergistic operation of these components, HoloDx achieves enhanced interpretability while maintaining precise knowledge-data alignment. 
Evaluations on five AD datasets demonstrate that HoloDx outperforms state-of-the-art methods, achieving superior diagnostic accuracy and strong generalization across diverse cohorts. The source code is released at \href{https://github.com/Qybc/HoloDx}{https://github.com/Qybc/HoloDx}.

\end{abstract}

\begin{IEEEkeywords}
Multimodal Alzheimer's Disease Diagnosis, Knowledge- and Data-Driven Framework, Large Language Models, Computer-Aided Diagnosis
\end{IEEEkeywords}

\begin{figure}[htbp]
        \centering
        \includegraphics[width=\linewidth]{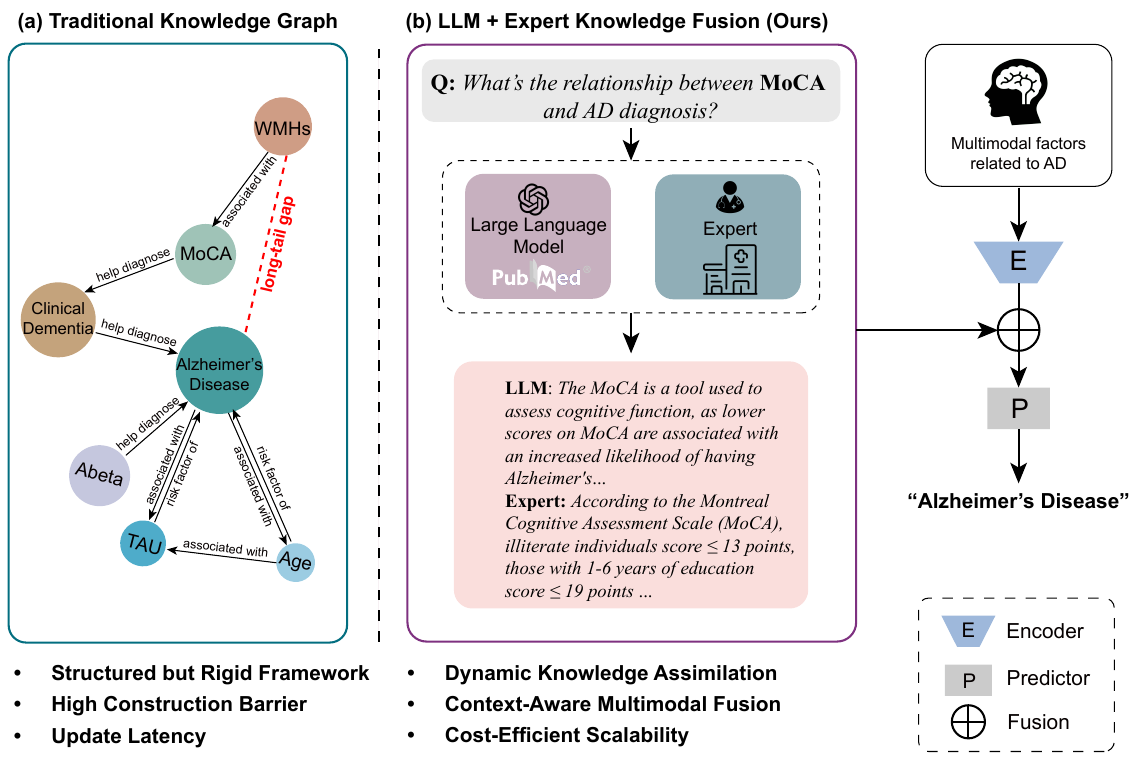}
        \caption{ 
        Illustration of our proposed computer-aided holistic diagnosis (HoloDx) system. (A) Traditional methods rely on predefined knowledge graphs that are limited by manual curation and slow updates. (B) Our framework combines large language model (LLM)-derived general medical knowledge with expert knowledge from hospitals, which is injected into a multimodal data-driven system to enable adaptive diagnostics.
}
		\label{fig:abstract}
\end{figure}

\section{Introduction}

Alzheimer's disease (AD) is a progressive neurodegenerative disorder that leads to memory loss, cognitive decline, and ultimately, loss of independence. Accurate diagnosis of AD is crucial for implementing appropriate treatment strategies, planning for future care, and providing support to patients and their families. Recent advancements~\cite{vanDyck2022LecanemabIE,Sims2023DonanemabIE,heneka2024passive,yi2024passive} in passive immunotherapies, such as Lecanemab and Donanemab, have shown promise in targeting amyloid-beta plaques, a hallmark of AD pathology. However, diagnosing AD, especially in its early stages, remains challenging due to subtle clinical symptoms and overlap with normal aging processes. Traditional diagnostic approaches, often relying on unimodal data such as neuroimaging or clinical assessments, struggle to capture these nuanced changes~\cite{jang2022m3t,dosovitskiy2020image}. To address these limitations, multimodal learning has gained attention for its ability to integrate diverse data sources, such as neuroimaging, clinical records, and genetic information, offering a more holistic perspective on AD's early pathological changes~\cite{chen2024alifuse,zhou2023transformer,yu2024transformer,chen2024smart}.
Despite the promise of data-driven approaches, significant challenges remain: 1) data bias, which focuses on prevalent patterns while often neglecting rare and atypical cases, and 2) a lack of interpretability, as many models function as opaque black boxes, making it difficult to trace causal relationships and ensure reliability. These limitations hinder the consistency, robustness, and trustworthiness of existing computer-aided diagnosis (CAD) systems.
Addressing these shortcomings requires a synergistic paradigm, as shown in Fig.~\ref{fig:abstract}, which integrates the strengths of multimodal data-driven models with domain knowledge, enhancing interpretability, reliability, and generalizability in the context of AD and its early diagnosis.

To effectively integrate data and domain knowledge, current methods have explored incorporating domain expertise into multimodal diagnostic systems. Many of these methods rely on predefined medical ontologies, such as knowledge graphs, to embed expert insights into the diagnostic process~\cite{ma2018kame}. For example, the Alzheimer’s Disease Knowledge Graph (ADKG)~\cite{yang2024alzheimer} encodes relationships between key concepts using curated medical abstracts. 
While these methods provide a structured representation of domain knowledge, they face significant limitations. Their static knowledge bases, like ADKG, make it difficult to adapt to individual patient data or changing clinical conditions, restricting their capacity to offer personalized insights. Moreover, despite attempts to compile extensive resources, these graphs typically represent only a portion of the ever-expanding and evolving body of medical knowledge, limiting their comprehensiveness. Also, integrating large-scale but mismatched knowledge could lead to redundancy and inconsistencies, making cross-modality alignment more difficult and compromising diagnostic reliability. These challenges highlight the need for innovative frameworks capable of adaptively incorporating coherent and personalized domain knowledge while maintaining synergy across modalities. 

Large language models (LLMs) have emerged as a promising tool for knowledge-driven medical diagnosis, offering expert-like reasoning and vast general knowledge. For example, HuatuoGPT-V~\cite{chen2024huatuogpt}, a multimodal model designed for medical consultation, demonstrates this potential. By leveraging extensive domain knowledge from sources like PubMed, LLMs can convey general medical facts through natural language interactions across multimodal data types. In addition to broad knowledge, LLMs can provide personalized insights by tailoring general information to individual patient contexts, improving clinical applicability and prediction accuracy. Also, expert knowledge is crucial for more context-specific interpretations of complex cases, filling the gaps left by LLMs' generalized approach. To combine these complementary sources, we propose merging LLM knowledge with domain expertise to create a unified knowledge bank, ensuring consistency, reducing redundancy, and encoding context-relevant information for more accurate and personalized diagnosis.

\begin{figure*}[htbp]
        \centering
        \includegraphics[width=\linewidth]{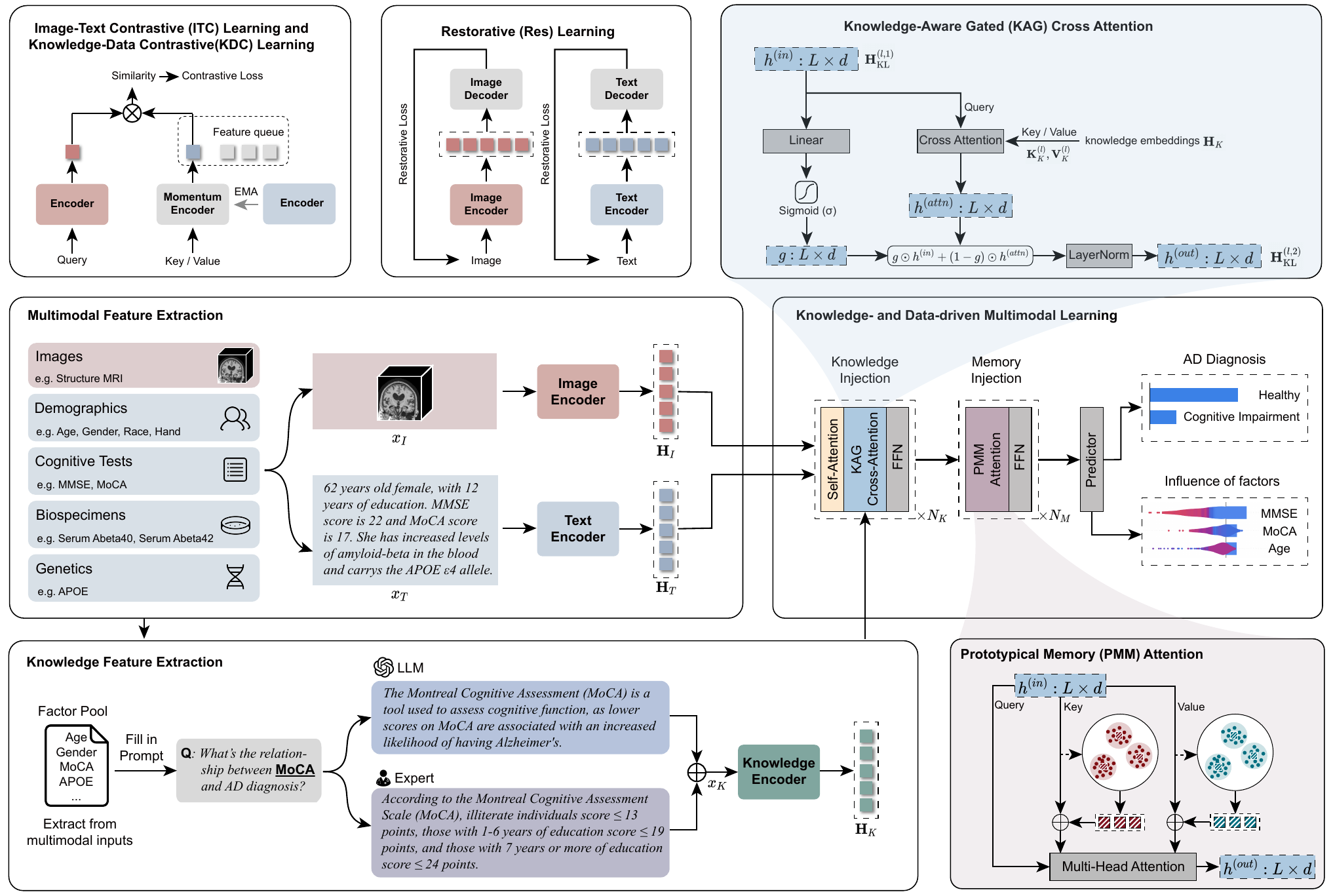} 
	\caption{Overview of our proposed HoloDx model. Multimodal data and domain knowledge serve as inputs, with multimodal features processed through the knowledge and memory injection modules, followed by a prediction layer for final diagnosis.}
		\label{fig:overview}
\end{figure*}

Building on the knowledge bank, we propose HoloDx, a novel knowledge- and data-driven CAD framework, as shown in Fig.~\ref{fig:overview}. HoloDx integrates multimodal data sources, such as imaging, cognitive tests, biospecimens, and genetic information, to provide accurate and universally applicable clinical assessments. By combining general insights from LLMs with specialized expertise from domain professionals, the system ensures reliable decision-making. To integrate this knowledge effectively, we design a knowledge injection module based on knowledge-aware gated cross-attention, enriching data samples with domain-specific insights and aligning visual and textual features. Additionally, a memory injection module with the proposed prototypical memory attention dynamically collects prior patient experience, mimicking expert knowledge accumulation and improving adaptability and performance.

Overall, our contributions are summarized as follows:
\begin{itemize}
    \item \textbf{A novel Dual-Driven CAD framework}: We propose HoloDx, a knowledge- and data-driven model designed to improve multimodal diagnosis of Alzheimer's disease and its early stages. By leveraging dynamic and adaptive domain knowledge learning, HoloDx achieves superior diagnostic accuracy, consistently outperforming state-of-the-art baselines across five datasets.
    \item \textbf{Injection of domain knowledge and dynamic prototypical memory}: We design a knowledge injection module to incorporate dynamic domain knowledge from medical LLMs and experts, ensuring informed multimodal representations. A dynamic prototypical memory mechanism further enables the model to adapt and refine its understanding by leveraging knowledge from past cases for improved performance.

\end{itemize}

\section{Related work}
\subsection{Multimodal CAD}

The integration of diverse modalities of medical data has become a prominent approach in advancing medical diagnosis~\cite{zhou2023integrating,xie2023orbital,xiao2023csifnet,ju2023m2}, enabling machine learning models to identify disease patterns more effectively than relying on single modalities. For example, Kim et al.~\cite{kim2023heterogeneous} develop a heterogeneous graph learning method to fuse multimodal medical data efficiently. Similarly, Zhou et al.~\cite{zhou2023transformer} introduce the IRENE model, which employs Transformer-based architectures to combine modality-specific low-level embeddings for diagnostic purposes. Multimodal integration has also succeeded in diagnosing conditions such as breast cancer, where combining imaging and genomic information has improved diagnostic outcomes~\cite{guez2022development}. 
Chen et al.~\cite{chen2024alifuse} propose a transformer-based framework for aligning and fusing multimodal medical data to classify Alzheimer’s disease.
These efforts highlight the growing potential of multimodal CAD systems to improve diagnostic precision and reliability. However, these approaches often rely solely on data-driven techniques, which limits their ability to handle complex medical scenarios where domain expertise is crucial. Without integrating domain knowledge, such models may fail to resolve conflicts between modalities or interpret rare and ambiguous cases effectively. This gap underscores the need for combining multimodal data integration with domain knowledge to enhance interpretability, consistency, and diagnostic reliability.

\subsection{Knowledge-driven Multimodal CAD}

The integration of domain knowledge has become a cornerstone strategy in healthcare to enhance patient phenotyping and clinical prediction accuracy. By encoding medical expertise into computational frameworks, models gain improved capacity to prioritize clinically relevant biomarkers and diagnostic patterns. Notable implementations include KAME~\cite{ma2018kame}, which employs attention-weighted medical knowledge graphs to model multi-disease correlations across clinical encounters. Subsequent architectures like LDAM~\cite{niu2021label}, CAML~\cite{mullenbach2018explainable}, and LERP~\cite{niu2021label2} further advance this paradigm through cross-attention mechanisms that align laboratory findings with disease-specific risk profiles. Recent innovations such as EHR-KnowGen~\cite{niu2024ehr} demonstrate enhanced diagnostic capability through multi-granular knowledge fusion, combining fine-grained symptom ontologies with coarse-grained treatment pathways.
Despite these advancements, current methodologies exhibit three fundamental constraints: (1) Static knowledge representations (e.g., ICD-based hierarchies) fail to adapt to patient-specific clinical trajectories; (2) Fixed graph structures inadequately model multimodal interactions between neuroimaging biomarkers, genetic markers, and longitudinal cognitive assessments; (3) Conventional attention mechanisms impose artificial sparsity constraints, hindering discovery of nonlinear biomarker interactions critical in complex disorders like Alzheimer's disease.

Our approach bridges this gap by dynamically enriching the knowledge bank with general medical insights derived from large language models (LLMs) and specific knowledge contributed by domain experts, tailored to the factors associated with each subject. This ensures that multimodal data is augmented with both broad and context-specific representations, enabling our method to provide more precise and interpretable predictions, effectively addressing key limitations in existing knowledge-driven frameworks.

\section{Methodology}

\begin{table}[t]
\begin{center}
\scriptsize
\caption{Dataset and factor details used in our experiments.}
\label{tab:factors}
\begin{tabular}{p{16pt}p{12pt}p{182pt}}
\toprule
\textbf{Dataset} & \textbf{Access} & \textbf{Factors} \\
\cmidrule(r){1-3}
\multirow{10}{*}{ADNI} & \multirow{10}{*}{Public} & \textbf{Image}: Hippocampus, Entorhinal Cortex, Amygdala, Thalamus, Cerebral White Matter, Cerebral Cortex, Lateral Ventricle, Cerebellum White Matter,
Cerebellum Cortex, Brain Stem, Choroid Plexus, Caudate, Putamen, Pallidum, Ventricle, Vessel \\
& & \textbf{Demographic}: Age, Gender, Education Year, Hand, Racial, Medical History (Neurologic, Cardiovascular, etc., 17 in total) \\
& & \textbf{Cognitive Test}: MMSE, MoCA, LDELTOTAL \\
& & \textbf{Biospecimen}: Thyroid Stim Hormone, Vitamin B12, Platelets, RBC, WBC, Serum Glucose, Urea Nitrogen, etc. (37 in total) \\
& & \textbf{Genetic}: APOE \\
\cmidrule(r){1-3}
\multirow{7}{*}{AIBL} & \multirow{7}{*}{Public} & \textbf{Image}: Same as ADNI \\
& & \textbf{Demographic}: Age, Gender, Medical History (Neurologic, Cardiovascular, etc., 9 in total) \\
& & \textbf{Cognitive Test}: MMSE, LIMMTOTAL, LDELTOTAL \\
& & \textbf{Biospecimen}: Thyroid Stim Hormone, Vitamin B12, Platelets, RBC, WBC, Urea Nitrogen, etc. (12 in total) \\
& & \textbf{Genetic}: APOE \\
\cmidrule(r){1-3}
\multirow{7}{*}{RENJI} & \multirow{7}{*}{Private} & \textbf{Image}: Same as ADNI \\
& & \textbf{Demographic}: Age, Gender, Education Year, Height, Weight, BMI \\
& & \textbf{Cognitive Test}: MMSE, MoCA, AVLT, GDS \\
& & \textbf{Biospecimen}: WBC, RBC, Serum Abeta40, Serum Abeta42, Serum pTau181, Serum pTau217, etc. (48 in total) \\
& & \textbf{Genetic}: APOE \\
\cmidrule(r){1-3}
\multirow{2}{*}{OASIS} & \multirow{2}{*}{Public} & \textbf{Image}: Same as ADNI; \textbf{others}: Gender, Age, Hand, Education Year, Socioeconomic Status, MMSE, CDR \\
\cmidrule(r){1-3}
MIRIAD & Public & \textbf{Image}: Same as ADNI; \textbf{others}: Gender, Age, MMSE \\
\bottomrule
\end{tabular}
\end{center}
\end{table}

\subsection{Knowledge and Data Collection}
\subsubsection{Factor Extraction}

Unlike existing methods~\cite{ma2018kame,niu2021label}, which primarily focus on general textual knowledge, our approach introduces fine-grained domain knowledge tailored to the factor level and extends factor extraction to a multimodal setting. We define factors as key elements or features derived from multimodal data that are critical for diagnosis, as listed in Table~\ref{tab:factors}. These factors are categorized into two main types: imaging factors and non-imaging factors:
\begin{itemize}
    \item Imaging factors ($x_I$): Derived from brain MRI scans using a pre-trained segmentation model~\cite{zhang2023pretrain}, imaging factors include critical brain structures, e.g., the hippocampus, ventricles, cortex, etc. Fine-grained domain knowledge is injected at this level, providing insights into how specific anatomical structures relate to AD diagnosis.
    \item Non-imaging factors ($x_T$): Extracted from clinical data, non-imaging factors include demographics (e.g., age, gender, education), cognitive test scores (e.g., MMSE, MoCA), biospecimen data (e.g., Thyroid Stim Hormone, Vitamin B12), and genetic information (e.g., APOE). Domain knowledge is incorporated into each non-imaging factor, linking specific biomarkers or cognitive metrics to disease diagnosis and clinical interpretations.
\end{itemize}

The extracted imaging and non-imaging factors are fused to form a set of multimodal matched factors ($F$), aligning diverse modalities into a unified representation enriched with domain knowledge. By integrating fine-grained insights directly at the factor level, our method enables more precise, interpretable, and clinically relevant predictions. 

\begin{table}[t]
\begin{center}
\scriptsize
\caption{Knowledge from LLMs and Experts for sampled factors.
}
\label{tab:knowledge}
\begin{tabular}{p{20pt}p{200pt}}
\toprule
\textbf{Factor} & \textbf{Knowledge}  \\
\cmidrule(r){1-2}
Serum pTau217 & \textbf{LLM}: Serum pTau217 is a biomarker that can be used to aid in the diagnosis of Alzheimer's disease, as higher levels are associated with increased risk and progression of AD. \\
& \textbf{Expert}: The increase of pTau217 in the serum of AD patients is usually early and significant, and its diagnostic power is generally superior to that of Abeta42/40 and pTau181. Therefore, pTau217 holds high clinical value in the early diagnosis of Alzheimer's disease. \\
\cmidrule(r){1-2}
MMSE & \textbf{LLM}: The MMSE score serves as a crucial instrument in evaluating cognitive abilities, helping to detect those potentially afflicted with AD. Typically, scores beneath the threshold of 24 on the MMSE suggest potential cognitive deficits, possibly linked to AD. \\
& \textbf{Expert}: The MMSE score criteria for suspecting cognitive impairment are as follows: $\leq$ 19 for the illiterate, $\leq$ 22 for those with a primary school background, and $\leq$ 26 for individuals with junior high school education or higher. \\
\cmidrule(r){1-2}
APOE & \textbf{LLM}: APOE is a genetic marker that increases the risk of developing Alzheimer's disease, with certain alleles being more strongly associated than others. \\
& \textbf{Expert}: The APOE e4 allele is the strongest genetic risk factor for sporadic Alzheimer's disease. Compared to the most common APOE e3 allele, individuals with one APOE e4 allele have about 3.7 times the risk of developing Alzheimer's disease, and those with two APOE e4 alleles face up to a 12-fold increased risk. \\
\bottomrule
\end{tabular}
\end{center}
\end{table}

\begin{table}[t]
\begin{center}
\scriptsize
\caption{Factor details provided by experts.
}
\label{tab:expert}
\begin{tabular}{p{20pt}p{200pt}}
\toprule
\textbf{Factors} & Serum Abeta40, Serum Abeta42, Serum pTau181, Serum pTau217, Serum Abeta42/40, MMSE, MoCA, AVLT, GDS, BMI, Age, Education, Gender, APOE, Hippocampus, Cerebral Cortex, Amygdala \\
\bottomrule
\end{tabular}
\end{center}
\end{table}

\subsubsection{Factor Knowledge Generation}

Based on the extracted factors, we enhance AD diagnosis by generating domain knowledge tailored to them. Our knowledge generator utilizes the extensive capabilities of LLMs to produce a comprehensive set of general insights. Using a text-based prompting strategy, we query the LLM with inputs specifically designed to elicit detailed and relevant information for each factor. For example, to obtain general knowledge about a factor’s relationship with AD diagnosis, we prompt the LLM with:
{\begin{verbatim}
Q: Please describe the relationship between 
   {factor} and AD diagnosis.
\end{verbatim}}
\noindent

Here, $\{\text{factor}\}$ refers to a specific factor $f \in F$, as listed in Table~\ref{tab:factors}. Examples include the Mini-Mental State Examination (MMSE) scores collected from cognitive tests. The generated responses offer comprehensive descriptions of each factor and its relevance to diagnosing AD, as presented in the LLM knowledge column of Table~\ref{tab:knowledge}.

While LLMs provide general insights for each factor, clinical precision often requires nuanced interpretations grounded in medical practice or specialized research findings. To ensure context-specific relevance, we integrate the broad knowledge generated by LLMs with domain expertise contributed by medical professionals. Specifically, AD experts enrich the interpretation of critical factors (see Table~\ref{tab:expert}) in our experiments. Their contributions are sampled in the expert knowledge column of Table~\ref{tab:knowledge}, showcasing the synergy between LLM-generated insights and expert-provided contextual knowledge.

\subsubsection{Knowledge Bank} We first construct an offline knowledge bank to effectively manage the integration of general knowledge generated by LLMs and specific knowledge provided by experts. This bank stores knowledge for pre-identified factors, ensuring the model has immediate access to comprehensive domain insights during training and inference. When new factors arise, the LLM is queried to generate relevant general knowledge, which is then combined with expert-provided insights if available and added to the knowledge bank. By dynamically incorporating new information, the knowledge bank not only ensures the system remains current with evolving data but also enhances the model’s adaptability and scalability, enabling it to handle diverse and expanding diagnostic scenarios with precision.

\subsubsection{Multimodal Data Collection}
The multimodal inputs consist of imaging data, such as structural MRI scans, and non-imaging data, including demographic information, cognitive test results, biospecimen analyses, and genetic profiles. For non-imaging clinical data, such as the MMSE scores, we argue that their numerical values and categorical information are more semantically meaningful when presented in textual descriptions. To this end, we transform raw data entries into descriptive text representations. For example, an MMSE score of 29 is represented as “The MMSE score is 29”. As illustrated in Fig.~\ref{fig:overview}, all non-imaging data are systematically converted into text representations to serve as inputs to our model.

\subsection{Multimodal and Knowledge Feature Extraction}
With the factor-specific knowledge generated, we proceed to multimodal feature extraction, where image, text, and knowledge representations are systematically processed to capture meaningful and context-aware features for AD diagnosis.

\noindent
\textbf{Unimodal Encoder.}
We employ unimodal encoders to process both visual inputs $x_I$ and textual inputs $x_T$. For the visual modality, a 3D Vision Transformer~\cite{dosovitskiy2020image} is used to encode the input images into vector representations, generating an image embedding that includes a [CLS] token to represent the global image feature. For the textual modality, we use a Longformer Transformer~\cite{beltagy2020longformer}, which maps long text representations into feature embeddings. Similarly, a [CLS] token is appended to the beginning of the text input to summarize the whole sentence representation. Each unimodal encoder consists of transformer blocks comprising self-attention (SA) layers and feed-forward networks (FFNs).

\noindent
\textbf{Unimodal Decoder.}
To preserve the fine-grained understanding of image and text modalities, we incorporate a unimodal decoder following each unimodal encoder. This lightweight decoder consists of six transformer blocks with an embedding dimension of 768. Each transformer block includes an SA layer and an FFN. For the image modality, a linear projection layer reconstructs the image representation; similarly, another linear projection layer reconstructs the text representation.

\noindent
\textbf{Knowledge Encoder.}
The textual knowledge set, denoted as $x_K$, comprises $n_K$ factors that encapsulate the aggregated knowledge, which is encoded using a textual knowledge encoder. For consistency and to enable the extraction of robust features within a unified feature space, the knowledge encoder is designed to share weights with the textual encoder. The resulting knowledge embeddings function as the keys and values with the cross-attention layer of the subsequent knowledge injection module, facilitating the seamless integration of domain knowledge into the model.

\subsection{Knowledge Injection Module}
Given derived visual embeddings $\mathbf{H}_I$, textual embeddings $\mathbf{H}_T$, and knowledge embeddings $\mathbf{H}_K$, we address the critical challenge of effectively integrating domain knowledge with multimodal data. Our proposed knowledge injection module employs a gated cross-attention mechanism to dynamically fuse clinically relevant information while suppressing noise from irrelevant knowledge elements.

First, we concatenate visual and textual embeddings to construct joint queries, keys, and values:
\begin{equation}
\mathbf{Q}_D = \mathbf{K}_D = \mathbf{V}_D = [\mathbf{H}_I; \mathbf{H}_T].
\end{equation}
These multimodal embeddings initially undergo processing through self-attention (SA) mechanisms to systematically model interactions between modalities:
\begin{equation}
\begin{split}
& \mathbf{H}_{\text{KL}}^{(l,1)} = \text{LN}\left(\text{SA}(\mathbf{Q}_{D}^{(l)}, \mathbf{K}_{D}^{(l)}, \mathbf{V}_{D}^{(l)}) + \mathbf{Q}_{D}^{(l)}\right),
\end{split}
\end{equation}
where the SA layer enables dynamic reweighting of modality-specific features through learned attention patterns.

The contextualized representations $\mathbf{H}_{\text{KL}}^{(l,1)}$ then interact with domain knowledge via the Knowledge-Aware Gated (KAG) mechanism. Using $\mathbf{H}_{\text{KL}}^{(l,1)}$ as queries, KAG retrieves and injects relevant knowledge elements from $\mathbf{H}_K$ (serving as keys/values):
\begin{equation}
\begin{split}
& \mathbf{H}_{\text{KL}}^{(l,2)} = \text{KAG}\left(\mathbf{H}_{\text{KL}}^{(l,1)}, \mathbf{K}_{K}^{(l)}, \mathbf{V}_{K}^{(l)}\right),
\end{split}
\end{equation}
establishing explicit connections between data-driven patterns and clinical knowledge bases. 

Finally, the full layer transformation integrates these components through residual learning:
\begin{equation}
\begin{split}
& \mathbf{H}_{\text{KL}}^{(l+1)} = \text{LN}\left(\text{FFN}(\mathbf{H}_{\text{KL}}^{(l,2)}) + \mathbf{H}_{\text{KL}}^{(l,2)}\right).
\end{split}
\end{equation}

As detailed in Fig. 2, the KAG operator implements parametric feature gating:
\begin{equation}
\begin{split}
& \text{g} = \sigma(W_\text{g} \mathbf{H}_{\text{KL}}^{(l,1)} + b_\text{g}), \\
& \mathbf{H}^{(attn)} = \text{CA}(\mathbf{H}_{\text{KL}}^{(l,1)}, \mathbf{K}_{K}^{(l)}, \mathbf{V}_{K}^{(l)}), \\
& \mathbf{H}_{\text{KL}}^{(l,2)} = \text{LN}(\text{g} \odot \mathbf{H}_{\text{KL}}^{(l,1)} + (1-\text{g}) \odot \mathbf{H}^{(attn)}),
\end{split}
\end{equation}
where $W_\text{g} \in \mathbb{R}^{d \times d}$ and $b_\text{g} \in \mathbb{R}^d$ are learnable parameters, $\sigma$ denotes the sigmoid activation, and CA represents multi-head cross-attention.

The final patient representation aggregates modality-specific information through the concatenation of [CLS] tokens:
\begin{equation}
    \mathbf{h} = [\mathbf{h}_I;\mathbf{h}_T;\mathbf{h}_K],
\end{equation}
where $\mathbf{h}_I$, $\mathbf{h}_T$, and $\mathbf{h}_K$ correspond to the [CLS] tokens of visual, textual, and knowledge embeddings, respectively.

\subsection{Memory Injection Module}

In clinical practice, physicians manage diverse patient populations with unique medical histories and clinical profiles. While accumulated experience enhances expertise, conventional methods often struggle to systematically harness insights from growing patient datasets. To address this gap, we propose an innovative memory injection module integrated into our framework, designed as a dynamic memory repository. This module continuously adapts to evolving clinical contexts through collaborative interaction with knowledge injection module, enabling efficient storage and retrieval of insights from historical patient data.

To achieve this, we extend Barraco et al.'s prototypical memory framework for image captioning~\cite{barraco2023little}, adapting their memory vector architecture to develop our prototypical memory (PMM) attention mechanism. For a sequential input of mini-batches $[\mathbf{h}^0, \mathbf{h}^1, \cdots, \mathbf{h}^i, \cdots]$ containing training instance embeddings, we implement dual memory banks $\mathcal{B}_\mathbf{K}$ and $\mathcal{B}_\mathbf{V}$ at each network layer. These banks systematically archive keys and values from historical training samples, preserving information within a sliding temporal window spanning $T$ iterations. This design captures the evolving geometric structure of key-value relationships across training epochs. The active memory components at the $t$-th iteration are selected as:
\begin{equation}
\begin{aligned}
    \mathcal{B}_\mathbf{K} = [\mathbf{K}(\mathbf{h}^{(t-1)}),\mathbf{K}(\mathbf{h}^{(t-2)}),\cdots,\mathbf{K}(\mathbf{h}^{(t-T)})], \\
    \mathcal{B}_\mathbf{V} = [\mathbf{V}(\mathbf{h}^{(t-1)}),\mathbf{V}(\mathbf{h}^{(t-2)}),\cdots,\mathbf{V}(\mathbf{h}^{(t-T)})],
\end{aligned}
\end{equation}
where $\{\mathbf{K}(\mathbf{h}^i)\}$ and $\{\mathbf{V}(\mathbf{h}^i)\}$ represent complete key/value sets generated when processing mini-batch $\mathbf{h}^i$ through the network layer.

Prototype key vectors are constructed by clustering the key manifold and extracting representative centroids. Formally, for a predefined memory capacity $m$:
\begin{equation}
\mathcal{M}_{\mathbf{K}}=\left[\mathcal{M}_{\mathbf{K}}^{1}, \mathcal{M}_{\mathbf{K}}^{2}, \ldots, \mathcal{M}_{\mathbf{K}}^{m}\right]=\text{K-Means}_{m}\left(\mathcal{B}_{\mathbf{K}}\right),
\end{equation}
where $\text{K-Means}_{m}(\cdot)$ computes $m$ cluster centroids from the key memory bank. Value prototypes are subsequently derived through distance-weighted interpolation of values associated with geometrically proximate keys in the manifold. For each key prototype $\mathcal{M}_{\mathbf{K}}^i$, we compute its corresponding value prototype by aggregating the top-$k$ nearest neighbors in the key space:
\begin{equation}
    \begin{array}{c}
\mathcal{M}_{\mathbf{V}}=\left[\mathcal{M}_{\mathbf{V}}^{1}, \mathcal{M}_{\mathbf{V}}^{2}, \ldots, \mathcal{M}_{\mathbf{V}}^{m}\right], \\
\mathcal{M}_{\mathbf{V}}^{i}=\sum_{\left(\mathbf{K}^{j}, \mathbf{V}^{j}\right) \in \operatorname{top-k}\left(\mathcal{M}_{\mathbf{K}}^{i}\right)} e^{-d\left(\mathcal{M}_{\mathbf{K}}^{i}, \mathbf{K}^{j}\right)} \mathbf{V}^{j},
\end{array}
\end{equation}
where $\operatorname{top-k}\left(\mathcal{M}_{\mathbf{K}}^{i}\right)$ retrieves the $k$ closest (key, value) pairs to prototype $\mathcal{M}_{\mathbf{K}}^i$ based on distance metric $d(\cdot)$, ensuring contextual alignment between key clusters and their associated value representations.

To maintain memory bank relevance during training while minimizing computational overhead, the memory banks retain a fixed capacity of $T$ historical batches, with periodic refresh cycles that selectively incorporate recent key/value streams from network layers. Prototype vectors are regenerated from these updated banks and stored in $\mathcal{M}_{\mathbf{K}}$ and $\mathcal{M}_{\mathbf{V}}$. In implementation, this refresh process executes biweekly per epoch, preserving approximately two epochs' worth of contextual data. The overlapping memory contents between consecutive updates enhance training stability through gradual knowledge transition.

The forward propagation through layers $l$ and $l+1$ integrates layer normalization (LN) and residual pathways as:
\begin{equation}
\begin{split}
& \mathbf{H}_{\text{MEM}}^{(l,1)} = \text{LN}\left(\text{PMM}(\mathbf{Q}^{(l)}, \mathbf{K}^{(l)}, \mathbf{V}^{(l)}) + \mathbf{Q}^{(l)}\right), \\
& \mathbf{H}_{\text{MEM}}^{(l+1)} = \text{LN}\left(\text{FFN}(\mathbf{H}_{\text{MEM}}^{(l,1)}) + \mathbf{H}_{\text{MEM}}^{(l,1)}\right),
\end{split}
\end{equation}
where PMM attention synergizes memory-derived and input-specific patterns through:
\begin{equation}
\begin{aligned}
    & \text{PMM}(\mathbf{Q},\mathbf{K},\mathbf{V}) = \text{MHA}(\mathbf{Q}, \tilde{\mathbf{K}}, \tilde{\mathbf{V}}), \\
    & \tilde{\mathbf{K}} = [\mathcal{M}_{\mathbf{K}} ; \mathbf{K}], \quad \tilde{\mathbf{V}} = [\mathcal{M}_{\mathbf{V}} ; \mathbf{V}],
\end{aligned}
\end{equation}
with $\text{MHA}(\cdot,\cdot,\cdot)$ denoting standard multi-head attention.

The processed embeddings subsequently traverse a ReLU-activated multilayer perceptron (MLP)~\cite{nair2010rectified}, followed by softmax normalization~\cite{jang2016categorical} for final disease classification.

\subsection{Learning Objectives}

\subsubsection{Alignment}

We employ the image-text contrastive (ITC) loss~\cite{radford2021learning} to align image features $h_I$ and text features $h_T$ generated by the image and text encoders. The ITC loss $\mathcal{L}_{\text{itc}}$ maximizes similarity for positive image-text pairs while suppressing negative pairs, implemented through normalized cross-entropy over all pairwise similarities.
To enhance consistency between input data and domain knowledge in disease diagnosis, we propose the Knowledge-Data Contrastive loss (KDC). This mechanism maintains two dynamically updated queues: a \textit{Data queue} storing concatenated multimodal features $[h_I; h_T]$, and a \textit{Knowledge queue} containing domain embeddings $h_k$ generated by a dedicated knowledge encoder. 
The KDC loss $\mathcal{L}_{\text{kdc}}$ follows the contrastive paradigm of $\mathcal{L}_{\text{itc}}$, pulling matched knowledge-data pairs closer while distancing mismatched pairs. 
We implement momentum encoders updated via exponential moving average (EMA) following BLIP\cite{li2022blip} and ALBEF\cite{li2021align}. Specifically, the parameters of momentum image/text encoders ($\xi$) are updated as $\xi \leftarrow m_c \cdot \xi + (1 - m_c) \cdot \theta$, where $m_c=0.995$ is the momentum coefficient and $\theta$ denotes the parameters of the corresponding online encoders. 
All momentum encoders operate without gradient backpropagation.
This EMA-based strategy ensures feature consistency within the dynamically updated data and knowledge queues by decoupling momentum encoder optimization from the online model training. To prevent abrupt shifts in feature distribution, the queues are exclusively maintained using outputs from the momentum encoder.

\subsubsection{Restoration}
Our restorative learning module is designed to enhance the global semantic understanding by incorporating fine-grained visual and textual information. 
That is, the feature extraction is augmented by a reconstruction learning branch, which includes an image decoder to reconstruct the original image from the representation and minimizes the pixel-level distance between the original image $x_I$ and the reconstructed image $x_I'$:
$\mathcal{L}_{res}^I =  \mathbb {E}_{x_I} \; \mathcal{D}_I(x_I, x_I')
$,
where $\mathcal{D}_I(x_I, x_I')$ presents the distance function that measures similarity between $x_I$ and $x_I'$, e.g., Mean Square Error (MSE), or L1 norm. We use MSE following the common setting~\cite{he2022masked}.
For the textual component, we apply a similar approach. 
A text decoder is trained to minimize the token-level distance between the original text $x_T$ and the reconstructed text $x'_T$:
$\mathcal{L}_{res}^T = \mathbb {E}_{x_T} \; \mathcal{D}_T(x_T, x_T')
$,
where $\mathcal{D}_T(x_T, x_T')$ is the distance function measuring text similarity, such as the commonly-used cross-entropy loss.

\subsubsection{Classification}
The diagnostic classification prediction module leverages each subject's ground-truth (GT) labels for supervision. Let the disease prediction for the $i$-th subject be denoted as $\hat{y}^i$. Given the GT label $y^i$, the cross-entropy loss for disease prediction is defined as:
$\mathcal{L}_{cls} = - \frac{1}{N} \sum_{i=1}^{N} y^i \text{log} (\hat{y}^i)
$,
where $N$ is the total number of training subjects.

The total loss includes the following five terms:
\begin{equation}
    \mathcal{L} = \lambda_{al} (\mathcal{L}_{itc} + \mathcal{L}_{kdc}) + \lambda_{res} (\mathcal{L}_{res}^I + \mathcal{L}_{res}^T) + \lambda_{cls} \mathcal{L}_{cls},
\end{equation}
where $\lambda_{al}$, $\lambda_{res}$, and $\lambda_{cls}$ are constants balancing these terms. 

\begin{table}[t]
\scriptsize
\begin{center}
    \caption{Characteristics of participants of five datasets.
    }
    \label{tab:datasets}
    \begin{tabular}{p{20pt}p{16pt}p{40pt}p{12pt}p{20pt}p{20pt}p{22pt}}
    \toprule
        \textbf{Dataset} & \textbf{Type} & \textbf{\#Subjects(F/M)} & \textbf{\#MRIs} & \textbf{Age} & \textbf{MMSE} & \textbf{MoCA} \\
    \cmidrule(r){1-7}
        \multirow{3}*{ADNI} & NC & 724(407/317) & 3205 & 72.9$\pm$6.3 & 29.0$\pm$1.2 & 26.0$\pm$2.6 \\
        & MCI & 843(344/499) & 4115 & 72.9$\pm$7.4 & 27.5$\pm$2.2 & 23.5$\pm$3.2 \\
        & AD & 519(222/297) & 2019 & 74.3$\pm$7.3 & 21.6$\pm$4.7 & 16.6$\pm$5.2  \\
    \cmidrule(r){1-7}
        \multirow{3}*{AIBL} & NC & 361(159/202) & 708 & 73.8$\pm$6.6 & 28.8$\pm$1.7 & - \\
        & MCI & 91(48/43) & 149 & 76.2$\pm$7.0 & 27.1$\pm$2.0 & -  \\
        & AD & 81(31/50) & 140 & 75.2$\pm$8.1 & 19.8$\pm$6.5 & -  \\
    \cmidrule(r){1-7}
        \multirow{2}*{OASIS} & NC& 72(50/22) & 172 & 76.9$\pm$8.1 & 29.2$\pm$0.9 & - \\
        & AD& 64(28/36) & 135 & 76.2$\pm$6.9 & 24.4$\pm$4.5 & - \\
    \cmidrule(r){1-7}
        \multirow{2}*{MIRIAD} & NC & 30(14/16) & 42 & 68.1$\pm$1.2 & 29.5$\pm$1.0 & - \\
        & AD & 179(104/75) & 244 & 69.9$\pm$7.0 & 17.5$\pm$4.8 & - \\
        \cmidrule(r){1-7}
        \multirow{3}*{RENJI} & NC& 101(54/47) & 101 & 67.9$\pm$7.9 & 28.3$\pm$1.8 & 22.7$\pm$5.4 \\
        & MCI & 90(57/33) & 90 & 72.2$\pm$9.3 & 23.3$\pm$4.5 & 15.0$\pm$5.5\\
        & AD & 180(106/74) & 180 & 71.9$\pm$7.6 & 19.2$\pm$6.0 & 13.8$\pm$5.4\\
    \bottomrule
    \end{tabular}
  \end{center}
\end{table}

\begin{table*}[t]
\begin{center}
\scriptsize
\caption{Performance comparison (\%) on ADNI, AIBL, and RENJI datasets. The best results are in bold.}
\label{tab:results}
\begin{tabular}{llllllllll} 
\toprule
\multirow{2}*{Datasets} & \multirow{2}*{Methods}  & \multicolumn{4}{c}{NC vs. CI} & \multicolumn{4}{c}{NC vs. MCI} \\
\cmidrule(r){3-6} \cmidrule(r){7-10}
& & \textbf{ACC} & \textbf{AUC} & \textbf{SEN} & \textbf{SPE} & \textbf{ACC} & \textbf{AUC} & \textbf{SEN} & \textbf{SPE} \\

\cmidrule(r){1-10}
\multirow{11}*{ADNI} & 3D ResNet\cite{he2016deep}  & 72.00$\pm$14.85 & 52.63$\pm$8.86  & 9.71$\pm$12.20 & \textbf{95.56$\pm$8.89} & 67.69$\pm$7.53 & 59.61$\pm$11.04 & 32.33$\pm$24.58 & \textbf{86.89$\pm$13.01} \\
& 3D ViT~\cite{dosovitskiy2020image}  & 76.00$\pm$16.11 & 58.91$\pm$15.58 & 25.04$\pm$29.52 & 92.78$\pm$8.89 & 63.08$\pm$13.23 & 54.57$\pm$14.01 & 30.67$\pm$24.73 & 78.48$\pm$14.87 \\
& BERT~\cite{devlin2018bert} & 85.33$\pm$7.77 & 78.25$\pm$8.68  & 64.76$\pm$18.65 & 91.74$\pm$4.42 & 81.53$\pm$9.23 & 80.29$\pm$11.06 & 84.33$\pm$16.38 & 86.25$\pm$15.33 \\
& RoBerta~\cite{liu2019roberta} & 86.67$\pm$9.43 & 81.58$\pm$9.71  & 70.10$\pm$18.67 & 93.06$\pm$9.04 & 87.69$\pm$7.84 & 84.38$\pm$10.45 & 83.00$\pm$19.04 & 85.77$\pm$12.18 \\
& Longformer\cite{beltagy2020longformer}  & 88.00$\pm$8.84 & 86.54$\pm$10.04 & 79.61$\pm$17.03 & 93.46$\pm$9.70 & 86.15$\pm$5.75 & 85.83$\pm$7.58 & 80.33$\pm$16.75 & 81.33$\pm$11.26 \\
& GIT~\cite{wang2022git} & 80.89$\pm$4.08 & 75.15$\pm$4.11 &  81.75$\pm$4.92 & 80.61$\pm$4.60 & 84.11$\pm$4.78 & 71.05$\pm$5.24 &  83.12$\pm$5.33 & 52.00$\pm$11.27  \\
& IRENE~\cite{zhou2023transformer}  & 87.06$\pm$4.78 & 78.44$\pm$7.36 &  90.88$\pm$5.34 & 66.00$\pm$14.20 & 86.47$\pm$2.99 & 78.10$\pm$6.41 &  90.19$\pm$4.18 & 65.99$\pm$14.20 \\
& AD-Trans~\cite{yu2024transformer}  & 86.80$\pm$6.36 & 81.87$\pm$8.05 &  \textbf{93.57$\pm$1.43} & 70.17$\pm$15.53 & 81.32$\pm$4.41 & 75.66$\pm$4.19 &  65.83$\pm$5.89 & 85.49$\pm$4.26  \\
& Alifuse~\cite{chen2024alifuse} & 87.17$\pm$5.38 & 82.46$\pm$8.15 &  92.83$\pm$3.78 & 72.09$\pm$13.22 &  87.05$\pm$4.77 & 78.44$\pm$7.39 &  90.88$\pm$5.34 & 73.44$\pm$11.20 \\
\cmidrule(r){2-10}
\rowcolor{myblue}
\cellcolor{white} & HoloDx (ours)  & \textbf{93.33$\pm$5.96} & \textbf{91.83$\pm$7.64} & 88.67$\pm$15.72 & 95.00$\pm$9.99 & \textbf{92.82$\pm$2.99} & \textbf{90.09$\pm$5.01} & \textbf{96.60$\pm$2.11} & 83.59$\pm$9.74  \\ 
\cmidrule(r){1-10}

\multirow{11}*{AIBL} & 3D ResNet\cite{he2016deep}  & 65.88$\pm$7.58 & 55.17$\pm$9.12 &  70.68$\pm$7.97 & 39.67$\pm$14.31 & 60.59$\pm$8.84 & 58.45$\pm$14.49 &  61.56$\pm$8.09 & 55.33$\pm$26.72 \\
& 3D ViT~\cite{dosovitskiy2020image} & 67.18$\pm$5.23 & 55.75$\pm$7.89 &  82.07$\pm$6.65 & 29.42$\pm$15.61 & 66.15$\pm$3.77 & 54.46$\pm$7.20 &  81.31$\pm$5.46 & 27.60$\pm$16.27\\
& BERT~\cite{devlin2018bert}  & 86.67$\pm$2.51 & 86.62$\pm$2.13 &  86.57$\pm$4.18 & \textbf{86.65$\pm$4.77} & 85.29$\pm$4.16 & 71.77$\pm$4.99 &  91.55$\pm$4.90 & 52.00$\pm$11.27 \\
& RoBerta~\cite{liu2019roberta}  & 87.69$\pm$4.10 & 86.35$\pm$5.48 &  88.48$\pm$4.91 & 84.23$\pm$10.38 & 88.82$\pm$4.32 & 82.07$\pm$4.97 &  91.48$\pm$5.82 & 72.67$\pm$12.89 \\
& Longformer\cite{beltagy2020longformer} & 88.21$\pm$3.84 & 86.24$\pm$4.97 &  90.72$\pm$3.05 & 81.76$\pm$9.71 & 88.23$\pm$4.16 & 75.63$\pm$8.57 &  93.60$\pm$4.75 & 57.67$\pm$19.60\\
& GIT~\cite{wang2022git}  & 84.35$\pm$5.43 & 82.62$\pm$5.13 &  88.57$\pm$5.18 & 84.65$\pm$4.77 & 83.29$\pm$7.90 & 82.77$\pm$4.79 &  89.55$\pm$5.90 & 64.20$\pm$12.76  \\
& IRENE~\cite{zhou2023transformer}  & 84.10$\pm$6.36 & 71.70$\pm$8.52 &  93.20$\pm$2.16 & 44.19$\pm$16.55 & 84.29$\pm$5.22 & 81.53$\pm$8.07 &  84.62$\pm$6.28 & 67.00$\pm$13.24  \\
& AD-Trans~\cite{yu2024transformer} & 85.64$\pm$6.99 & 80.52$\pm$5.93 &  92.62$\pm$7.84 & 68.44$\pm$8.60 & 83.84$\pm$4.47 & 77.92$\pm$6.13 &  91.55$\pm$5.92 & 64.29$\pm$11.49 \\
& Alifuse~\cite{chen2024alifuse} & 86.67$\pm$4.41 & 83.88$\pm$6.27 &  90.01$\pm$3.73 & 77.74$\pm$9.73 & 86.15$\pm$5.28 & 77.99$\pm$7.56 &  87.31$\pm$7.08 & 64.33$\pm$12.45  \\

\cmidrule(r){2-10}      

\rowcolor{myblue}
\cellcolor{white} & HoloDx (ours)  & \textbf{91.80$\pm$3.83} & \textbf{89.38$\pm$5.32} & \textbf{96.60$\pm$2.11} & 80.56$\pm$5.27 & \textbf{91.28$\pm$3.07} & \textbf{87.95$\pm$4.93} & \textbf{94.92$\pm$2.99} & \textbf{80.97$\pm$10.80} \\ 
\cmidrule(r){1-10}

\multirow{11}*{RENJI} & 3D ResNet\cite{he2016deep}  & 62.67$\pm$6.79 & 53.99$\pm$4.44 &  15.67 $\pm$8.41 & 92.32$\pm$10.70 & 61.33$\pm$6.53 & 53.01$\pm$9.33 & 20.38$\pm$17.03 & 85.64$\pm$9.79 \\
& 3D ViT~\cite{dosovitskiy2020image} & 69.33$\pm$13.06 & 64.24$\pm$11.74 &  47.10$\pm$19.03 & 81.39$\pm$16.00 & 64.00$\pm$6.80 &  54.17$\pm$9.45 &  16.38$\pm$14.04 & 91.96$\pm$7.52 \\
& BERT~\cite{devlin2018bert}  & 86.67$\pm$11.16 & 86.07$\pm$11.49 &  82.62$\pm$14.43 & 89.52$\pm$12.20 & 82.67$\pm$13.06 &  82.07$\pm$14.85 &  80.48$\pm$19.54 & 83.66$\pm$12.19 \\
& RoBerta~\cite{liu2019roberta}  & 89.33$\pm$6.80 & 87.20$\pm$7.21 &  78.62$\pm$11.55 & \textbf{95.89$\pm$6.33} & 88.00$\pm$4.98 & 89.65$\pm$5.68 & 94.29$\pm$11.43 & 85.02$\pm$5.17 \\
& Longformer\cite{beltagy2020longformer} & 88.01$\pm$9.79 & 86.21$\pm$12.25 &  78.96$\pm$12.23 & 93.46$\pm$9.70 & 88.00$\pm$7.78 & 88.12$\pm$7.42 &  86.95$\pm$11.34 & 89.29$\pm$12.20 \\
& GIT~\cite{wang2022git}  & 81.33$\pm$14.24 & 76.48$\pm$14.36 & 59.62$\pm$17.24 & 93.33$\pm$13.33 & 80.00$\pm$11.93 & 83.25$\pm$10.31 & 90.48$\pm$3.13 & 76.02$\pm$16.28 \\
& IRENE~\cite{zhou2023transformer}  & 82.67$\pm$6.80 & 85.06$\pm$5.05 &  90.29$\pm$12.20 & 79.84$\pm$13.60 & 84.00$\pm$9.98 & 86.07$\pm$8.08 &  90.48$\pm$2.01 & 81.66$\pm$11.80 \\
& AD-Trans~\cite{yu2024transformer} & 83.84$\pm$7.53 & 80.10$\pm$10.78 &  88.73$\pm$1.42 & 71.43$\pm$11.29 & 82.83$\pm$5.27 & 75.72$\pm$6.01 &  92.96$\pm$1.12 & 57.14$\pm$14.76 \\
& Alifuse~\cite{chen2024alifuse} & 89.33$\pm$7.98 & 88.64$\pm$9.48 &  85.81$\pm$14.66 & 91.46$\pm$9.16 & 86.67$\pm$12.65 &  85.44$\pm$13.48 & 80.10$\pm$18.62 & 90.78$\pm$9.20  \\
\cmidrule(r){2-10}
\rowcolor{myblue}
\cellcolor{white} & HoloDx (ours)  & \textbf{94.67$\pm$4.99} & \textbf{94.80$\pm$5.17} & \textbf{93.81$\pm$7.62} & 95.78$\pm$5.18 & \textbf{93.33$\pm$5.96} & \textbf{93.37$\pm$6.35} & \textbf{90.95$\pm$11.70} & \textbf{95.78$\pm$5.18} \\ 
\bottomrule
\end{tabular}
\end{center}
\end{table*}

\begin{table*}[t]
\begin{center}
\scriptsize
\caption{Evaluations on the inter-center generalization of various approaches. All models are trained on the ADNI dataset and Evaluated on the RENJI, OASIS, and MIRIAD Dataset. The best results are in bold.}
\label{tab:inter-center}
\begin{tabular}{lllllllllllll} 
\toprule
\multirow{2}*{Methods}  & \multicolumn{4}{c}{NC vs. CI (RENJI)} & \multicolumn{4}{c}{NC vs. CI (OASIS)} & \multicolumn{4}{c}{NC vs. CI (MIRIAD)} \\
\cmidrule(r){2-5} \cmidrule(r){6-9} \cmidrule(r){10-13}
& ACC(\%) & AUC(\%) & SEN(\%) & SPE(\%) & ACC(\%) & AUC(\%) & SEN(\%) & SPE(\%) & ACC(\%) & AUC(\%) & SEN(\%) & SPE(\%) \\

\cmidrule(r){1-13}
3D ResNet\cite{he2016deep}  & 62.00 & 52.63  & 9.71 & 95.56 & 55.08 & 49.73 & 5.21 & 94.26 & 55.63 & 50.97 & 12.11 & 89.83 \\
3D ViT~\cite{dosovitskiy2020image}  & 65.33 & 54.00 & 10.00 & \textbf{98.01} & 64.46 & 50.58 & 8.83 & 89.33 & 65.39 & 53.60 & 13.08 & 94.12  \\
BERT~\cite{devlin2018bert} & 82.67 & 85.15 & 90.29 & 80.02 & 65.85 & 68.52 & 46.15 & 90.88 & 87.01 & 59.04 & 19.28 & 98.81 \\
RoBerta~\cite{liu2019roberta} & 84.00 & 80.54 & 67.76 & 93.33 & 48.70 & 53.27 & 14.68 & 91.86 & 75.79 & 69.52 & 60.39 & 78.65 \\
Longformer\cite{beltagy2020longformer}  & 88.00 & 86.52 & 82.48 & 90.56 & 47.48 & 52.89 & 7.27 & \textbf{98.51} & 88.07 & 59.64 & 19.28 & \textbf{99.97} \\
GIT~\cite{wang2022git} & 77.33 & 69.97 & 59.14 & 80.79 & 68.21 & 66.62 & 53.17 & 80.07 & 75.04 & 71.41 & 58.61 & 84.22 \\
IRENE~\cite{zhou2023transformer}  & 69.33 & 69.10 & 64.90 & 73.30 & 73.81 & 71.10 & 61.56 & 80.64 & 74.82 & 72.12 & 62.62 & 81.61  \\
AD-Trans~\cite{yu2024transformer} & 73.84 & 69.58 & 77.46 & 71.42 & 73.20 & 62.43 & \textbf{68.43} & 77.21 & 75.69 & 71.94 & 59.79 & 76.61 \\
Alifuse~\cite{chen2024alifuse} & 80.00 & 83.97 &  \textbf{97.14} & 70.80 & 75.18 & 71.92 & 60.53 & 83.32 & 76.19 & 74.91 & 70.43 & 79.39 \\
\cmidrule(r){1-13}
\rowcolor{myblue}
HoloDx (ours)  & \textbf{90.67} & \textbf{88.68} & 81.80 & 95.56 & \textbf{78.67} & \textbf{74.82} & 60.43 & 89.24 & \textbf{89.33} & \textbf{88.01} & \textbf{81.48} & 78.03 \\
\bottomrule
\end{tabular}
\end{center}
\end{table*}

\section{Experiments and Discussions}
\subsection{Datasets and Settings}

We evaluate our model on four public datasets (ADNI~\cite{petersen2010alzheimer}, AIBL~\cite{ellis2009australian}, OASIS~\cite{marcus2007open}, MIRIAD~\cite{malone2013miriad}) and one private dataset RENJI. The RENJI is derived from the \textit{Multimodal Imaging Biomarkers in Degenerative Dementia Cohort} (MIMI, ClinicalTrials number: NCT06534658\cite{NCT06534658}) at Renji Hospital, which is affiliated with the School of Medicine at the Shanghai Jiao Tong University.
Table~\ref{tab:datasets} presents the number of subjects and MRIs, as well as the distribution of demographic information, i.e., gender and age, and two key cognitive test values, i.e., MMSE and MoCA, across NC, MCI, and AD categories. Other non-imaging information is included in Table~\ref{tab:factors}.

Following established clinical guidelines~\cite{mckhann2011diagnosis,dubois2007research,jack2018nia}, we evaluate our model on two classification tasks. The first task distinguishes cognitively normal (NC) individuals from those with cognitive impairment (CI), including both mild cognitive impairment (MCI) and Alzheimer’s disease (AD). The second one focuses on differentiating NC from MCI, which is a critical stage for the early identification of AD. We split the ADNI, AIBL, and RENJI datasets {\it subject-wise} into training, validation, and test sets with proportions of 70\%, 10\%, and 20\%, respectively. 
All structural MRI scans underwent standardized preprocessing, including skull stripping~\cite{isensee2019automated} to remove non-brain tissues and intensity normalization to harmonize voxel value distributions across scanners. 
Also, we perform zero-shot tests on OASIS and MIRIAD datasets to assess inter-center generalization.

\noindent
\textbf{Baselines.}
To evaluate the performance of our model, we compare it with three groups of baselines:
(1) Image-only group.
This group includes two recent methods commonly used in image classification: 3D ResNet~\cite{he2016deep}, and 3D ViT~\cite{dosovitskiy2020image}.
(2) Text-only group.
We select three prominent text-only baselines: BERT~\cite{devlin2018bert}, Roberta~\cite{liu2019roberta}, and Longformer~\cite{beltagy2020longformer}.
(3) Multimodal group.
This group includes four recent transformer-based models that fuse multimodal information for classification: GIT~\cite{wang2022git}, IRENE~\cite{zhou2023transformer}, AD-Trans~\cite{yu2024transformer}, and Alifuse~\cite{chen2024alifuse}.

\noindent
\textbf{Implementation details}
For pre-processing, each image is resized and cropped to $128 \times 128 \times 128$, with a spacing of [1.0,1.0,1.0] and an image patch size of $16$. The maximum input text length of textual input is set to 512. Following common settings in~\cite{he2022masked}, we set all hidden feature sizes to 768 and the number of heads in multi-head attention (MHA) to 12. 
The number of layers in the unimodal encoder is set to 12, while the number of layers in the decoder is set to 6. The number of layers in the knowledge and the memory modules is set to 6. The iteration $T$ is 100.
The values for $\lambda_{al}, \lambda_{res}, \lambda_{cls}$ are empirically set to 1, 1, and 1, respectively. Our model is trained on four NVIDIA RTX 3090 GPUs with a batch size of $12$, using the AdamW optimizer with a learning rate of 2e-5. Evaluation metrics include diagnosis accuracy
(ACC), Area Under Curve (AUC), sensitivity (SEN), and specificity (SPE).

\begin{table}[t]
\begin{center}
    \caption{Ablation results (\%) on the AIBL test dataset. (Best in bold)}
    \label{tab:ablation}
    \begin{tabular}{lllll}
    \toprule
        \textbf{Task} & \textbf{ACC} & \textbf{AUC} & \textbf{SEN} & \textbf{SPE} \\
    \cmidrule(r){1-5}
        \textbf{(a) Loss Terms} \\
    \cmidrule(r){1-5}
        $\mathcal{L}_{\text{cls}}$ & 89.23 & 81.70 &  96.42 & 74.33  \\
        $\mathcal{L}_{\text{cls}}+\mathcal{L}_{\text{itc}}$ & 89.75 & 84.42 &  95.48 & 79.38   \\
        $\mathcal{L}_{\text{cls}}+\mathcal{L}_{\text{itc}}+\mathcal{L}_{\text{kdc}}$ & 91.09 & 84.80 &  95.83 & \textbf{81.76}   \\
        $\mathcal{L}_{\text{cls}}+\mathcal{L}_{\text{itc}}+\mathcal{L}_{\text{kdc}}+\mathcal{L}_{\text{res}}^I$ & 91.28 & 85.74 &  \textbf{98.54} & 72.92  \\
        $\mathcal{L}_{\text{cls}}+\mathcal{L}_{\text{itc}}+\mathcal{L}_{\text{kdc}}+\mathcal{L}_{\text{res}}^I + \mathcal{L}_{\text{res}}^T$ & \textbf{91.92} & \textbf{88.23} &  97.14 & 79.31  \\
    \cmidrule(r){1-5}
        \textbf{(b) Module Terms} \\
    \cmidrule(r){1-5}
        Backbone & 89.34 & 83.18 &  96.43 & 71.93\\
        Backbone + KL & 91.28 & 86.94 & 94.92 & \textbf{80.97} \\
        Backbone + KL( w/o KAG) + MEM & 88.72 & 80.87 & \textbf{99.28} & 62.45 \\
        Backbone + KL + MEM (ours) & \textbf{91.92} & \textbf{88.23} &  97.14 & 79.31  \\
    \cmidrule(r){1-5}
        \textbf{(c) Feature Terms} \\
    \cmidrule(r){1-5}
        Image & 71.58 & 54.20 & 95.71 & 11.67\\
        Clinical & 88.21 & 82.10 & 96.43 & 67.76\\
        Image + Knowledge & 75.76 & 60.64 & 97.12 & 24.14 \\
        Clinical + Knowledge & 89.54 & 84.77 & 96.43 & 73.11 \\
        Image + Clinical + Knowledge & \textbf{91.92} & \textbf{88.23} &  \textbf{97.14} & \textbf{79.31}\\
    \cmidrule(r){1-5}
        \textbf{(d) Cognitive Tests} \\
    \cmidrule(r){1-5}
        w/o & 79.80 & 79.43 & 80.28 & 78.57\\
        w/ & \textbf{91.92} & \textbf{88.23} &  \textbf{97.14} & \textbf{79.31}\\
        
    \bottomrule
    \end{tabular}
  \end{center}
\end{table}

\begin{figure*}[htbp]
    \centering
    \includegraphics[width=\linewidth]{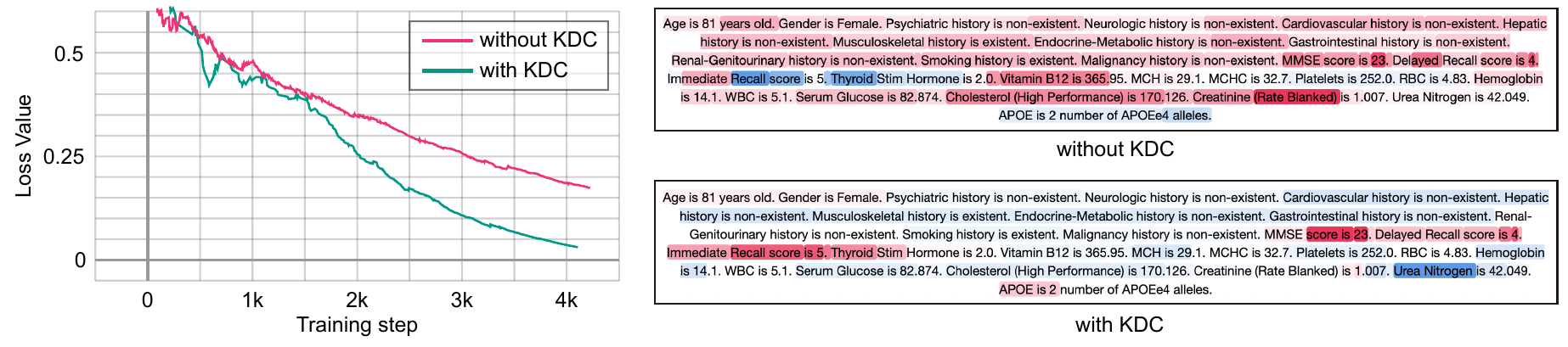}
    \caption{The impact of KDC loss on the AIBL dataset. Left: the convergence curve. Right: Shapley analysis of CI subjects from the test set. The color represents the contribution of each factor to the model's final prediction as CI. Red and blue represent positive and negative values, respectively.}
    \label{fig:kdc}
\end{figure*}

\begin{table*}[htbp]
\centering
\caption{Samples of the multimodal information in AIBL test dataset. (Top: CN; Bottom: CI).}
\label{tab:image_and_non-image}
\begin{tabular}{p{160pt}p{300pt}}
\toprule
\textbf{Image} & \textbf{Non-image} \\
\cmidrule{1-2}
\begin{minipage}[b]{0.6\columnwidth} \raisebox{-.9\height}{\includegraphics[width=\linewidth]{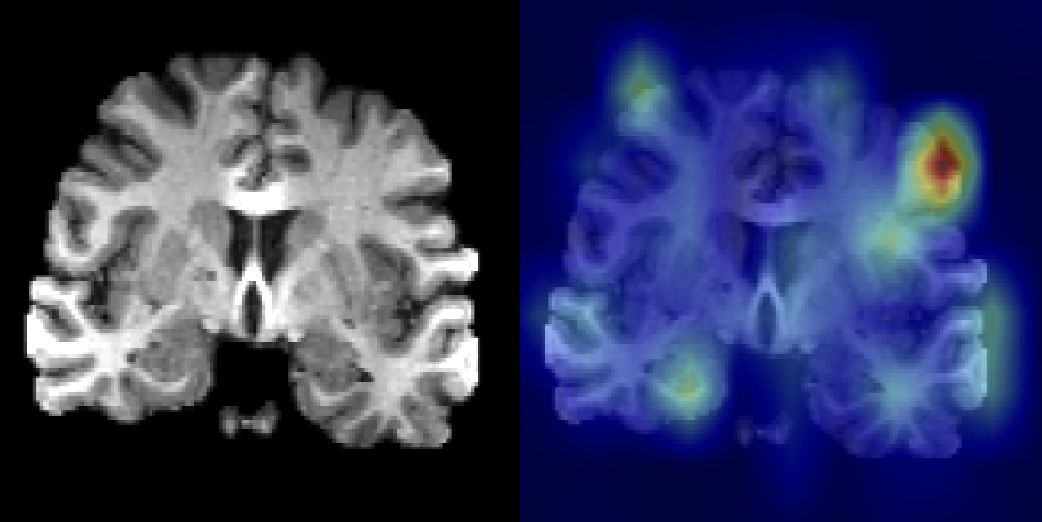}} \end{minipage} & Age is 70 years old. Gender is Male. Psychiatric history is non-existent. Neurologic history is non-existent. Cardiovascular history is non-existent. Hepatic history is non-existent. Musculoskeletal history is non-existent. Endocrine-Metabolic history is non-existent. Gastrointestinal history is non-existent. Renal-Genitourinary history is non-existent. Smoking history is existent. Malignancy history is non-existent. MMSE score is 27. Delayed Recall score is 5. Immediate Recall score is 6. Thyroid Stim Hormone is 1.9. Vitamin B12 is 528.594. MCH is 31.5. MCHC is 33.3. Platelets is 268.0. RBC is 4.13. Hemoglobin is 13.0. WBC is 5.7. Serum Glucose is 82.874. Cholesterol (High Performance) is 193.325. \\
\cmidrule{1-2}
\begin{minipage}[b]{0.6\columnwidth} \raisebox{-.9\height}{\includegraphics[width=\linewidth]{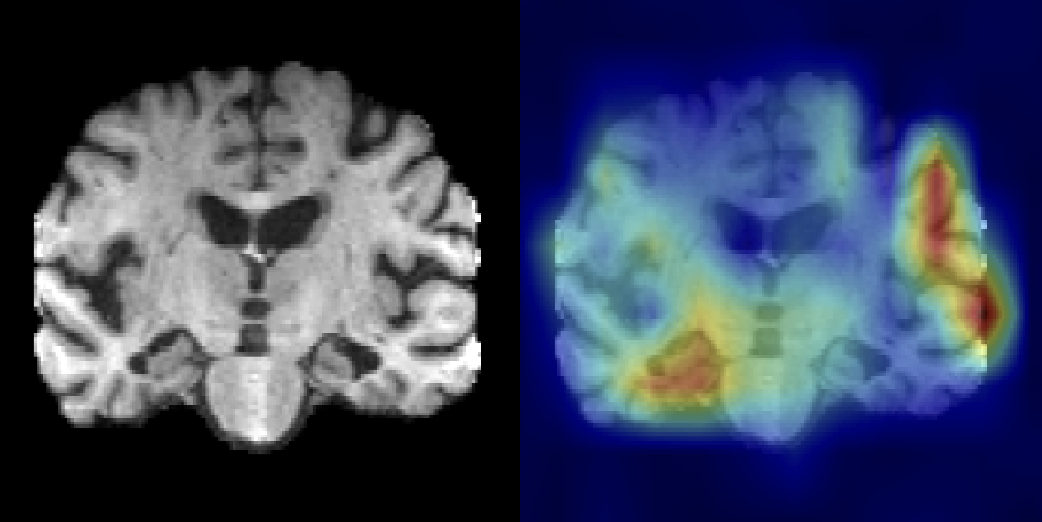}} \end{minipage}  & Age is 81 years old. Gender is Male. Psychiatric history is non-existent. Neurologic history is non-existent. Cardiovascular history is existent. Hepatic history is non-existent. Musculoskeletal history is existent. Endocrine-Metabolic history is non-existent. Gastrointestinal history is non-existent. Renal-Genitourinary history is non-existent. Smoking history is non-existent. Malignancy history is non-existent. MMSE score is 27. Delayed Recall score is 3. Immediate Recall score is 7. Thyroid Stim Hormone is 5.19. Vitamin B12 is 433.718. MCH is 32.3. MCHC is 34.7. Platelets is 252.0. RBC is 3.52. Hemoglobin is 11.4. WBC is 4.3. Serum Glucose is 90.08. Cholesterol (High Performance) is 201.058.  \\
\bottomrule
                                                 
\end{tabular}
\end{table*}

\subsection{Experimental Results}

\subsubsection{Intra-Diagnostic Classification} Table~\ref{tab:results} presents the results of our method on three datasets, i.e., ADNI, AIBL, and RENJI, comparing it with eight baselines. Our model outperforms image-only, text-only, and other multimodal models, underscoring the importance of integrating knowledge with imaging and textual data for medical diagnostics. Specifically, our results achieve accuracy and AUC values greater than 90\%, except for AIBL’s AUCs, which remain above 88\%. Notably, the text-only baselines outperform image-only models on the three datasets, as the text contains more comprehensive and directly relevant diagnostic information. Also, in multimodal approaches, the performance of models like IRENE and Alifuse is limited by challenges posed by long text inputs, resulting in performance not exceeding that of text-only unimodal models with a token length receptive field of 512. 

\subsubsection{Evaluation on Inter-Center Generalization}
To assess the generalization of our method, we conduct inter-center testing on the RENJI, OASIS, and MIRIAD datasets, with ADNI for training. Table~\ref{tab:inter-center} presents the performance of our HoloDx compared to baselines. 
Our method achieves the best performance among all models. However, the testing accuracy on OASIS is relatively lower, at 78.67\%, with an AUC of 74.82\%, partially due to the limited availability of non-imaging data and the model's strong reliance on MMSE scores (see the follow-up factor analysis). As shown in Table~\ref{tab:datasets}, the average MMSE score for AD in OASIS is relatively high compared to other datasets, making it more challenging to distinguish NC and AD. In contrast, HoloDx performs better on MIRIAD, as its MMSE distribution separates NC and AD more clearly.

\subsubsection{Ablation Studies}
We conduct three experiments to evaluate the contributions of each component in our framework. 

\textbf{(1) Impact of loss terms.} We evaluate the contribution of each loss, i.e., the classification loss $\mathcal{L}_{cls}$, two contrastive losses $\mathcal{L}_{itc}$ and $\mathcal{L}_{kdc}$, and the combined restorative loss $\mathcal{L}_{res}$ for images and texts. Incorporating all loss components is optimal for maximizing performance across all metrics, as reported in Table~\ref{tab:ablation}(a). Using only $\mathcal{L}_{cls}$, the model performance experiences a reduction of 6.5\% in AUC compared to the best-performing configuration. Adding $\mathcal{L}_{itc}$ and $\mathcal{L}_{kdc}$ alongside $\mathcal{L}_{cls}$ improves performance, increasing AUC by 3.1\%. This improvement highlights the benefits of incorporating contrastive learning between images and texts. 
The effectiveness of $\mathcal{L}_{\text{kdc}}$ lies in its unique combination of knowledge anchoring, noise suppression, and convergence acceleration. By aligning multimodal features $[h_I; h_T]$ with domain-specific knowledge embeddings $h_k$, $\mathcal{L}_{\text{kdc}}$ establishes a stable semantic reference frame grounded in clinical ontologies and medical guidelines, preventing feature drift during modality fusion. Unlike traditional image-text contrastive objectives that only align paired modalities, this loss ensures all features remain consistent with evidence-based medical knowledge. Moreover, the knowledge queue serves as a semantic filter, suppressing non-discriminative variations commonly found in medical data and promoting attention to clinically relevant biomarkers. This effect is supported by our feature shapley analysis as shown in Fig.~\ref{fig:kdc}.  Furthermore, $\mathcal{L}_{\text{kdc}}$ accelerates early-stage training by providing strong semantic gradients; as shown in Fig.~\ref{fig:kdc}, models without this component require over $1.5\times$ more iterations to achieve comparable representation quality. These combined benefits make $\mathcal{L}_{\text{kdc}}$ a catalytic component in our framework, significantly enhancing both robustness and learning efficiency. The highest diagnostic performance is achieved when the restoration loss $\mathcal{L}_{res}$ is included in training. Removing this type of losses on images and texts lead to a notable performance decline (3.43\% drop in AUC), confirming its essential role in preserving modality-specific features during cross-modal alignment. This highlights the importance of fine-grained reconstruction supervision in enhancing unimodal features quality and overall diagnostic effectiveness.

\textbf{(2) Impact of model components.}
To evaluate the effectiveness of individual components, we perform ablation experiments by selectively activating the Knowledge Module (KL), Memory Module (MEM), and Knowledge-aware Gated Attention (KAG), as reported in Table~\ref{tab:ablation}(b). Notably, introducing the Knowledge Module alone improves the baseline AUC by 3.76\%, and its combination with the Memory Module yields an additional 1.29\% gain. The full model achieves a total performance improvement of 5.05\% in AUC. Additionally, we replace the Knowledge-aware Gated Attention with standard additive attention. This substitution results in a significant performance drop, specifically 2.56\% in accuracy and 6.07\% in AUC, highlighting the critical role of gated attention in facilitating adaptive and effective feature fusion.

\textbf{(3) Impact of feature terms.}
We conduct an ablation study to evaluate the contributions of individual and partial modality combinations. As shown in Table~\ref{tab:ablation}(c), five configurations were compared: (1) image features only, (2) clinical data only (including demographics, cognitive assessments, biospecimens, and genetic information), (3) image + knowledge (general knowledge and expert rules), (4) clinical + knowledge, and (5) the full model incorporating all three feature terms.

The results reveal two key observations. First, clinical data provides broader contextual information than imaging alone. While the image-only model achieved an AUC of 54.20\% by capturing visual patterns, the clinical-only model reached 82.10\% AUC by leveraging trends from cognitive tests and historical medical data. The substantial 27.90\% performance gap indicates that clinical features encode systemic disease characteristics often missed by imaging. Second, the integration of clinical data with additional modalities yields the greatest performance gains. The ``clinical + knowledge" combination achieved 84.77\% AUC, while the full model integrating image, clinical, and knowledge modalities achieved the highest overall performance of 88.23\% AUC, demonstrating the complementary value of multi-modal information fusion.

\textbf{(4) Impact of cognitive scores and multimodal integration.}
To assess the potential dominance of cognitive scores such as MOCA and MMSE in the diagnostic process, we conduct an ablation experiment by excluding these features from the input. As shown in Table~\ref{tab:ablation}(e), removing MOCA and MMSE leads to a substantial performance drop: accuracy decreased by 12.12\% and AUC dropped by 8.8\% on the AIBL dataset. This result underscores the strong correlation between cognitive scores and AD diagnosis, aligning with their clinical use as standard screening tools. 

To further ensure robust decision-making, our model jointly leverages imaging data and interpretable mechanisms, preventing over-reliance on a single modality. Table~\ref{tab:image_and_non-image} presents representative cases in which patients with identical MMSE scores (i.e., MMSE = 27) receive different diagnostic outcomes based on structural MRI findings. For instance, patients showing pronounced hippocampal atrophy or ventricular enlargement are correctly identified as cognitively impaired (CI), while others with similar cognitive profiles but normal imaging features are classified as cognitively normal (CN). These results highlight the complementary nature of multimodal input, i.e., clinical data offer sensitive screening, while imaging data provide structural confirmation. As shown in Table~\ref{tab:ablation}(c), our multimodal design leads to a 6.13\% and 34.04\% AUC gain over clinical-only and image-only models, respectively. Such synergy reinforces the necessity of integrating both modalities for comprehensive AD assessment. 

\begin{table}[t]
\setlength{\tabcolsep}{3.5pt} 
\centering
\caption{Impact of Knowledge Type on model performance (\%) on AIBL and RENJI test sets.}
\label{tab:knowledge_type}
\begin{tabular}{lcccccccc}
\toprule
\multirow{2}{*}{\textbf{Type}} & \multicolumn{4}{c}{\textbf{AIBL}} & \multicolumn{4}{c}{\textbf{RENJI}} \\
\cmidrule(lr){2-5} \cmidrule(lr){6-9}
 & \textbf{ACC} & \textbf{AUC} & \textbf{SEN} & \textbf{SPE} & \textbf{ACC} & \textbf{AUC} & \textbf{SEN} & \textbf{SPE} \\
\midrule
LLM          & 90.25 & 84.67 & 97.73 & 71.59 & 91.20 & 90.67 & 83.99 & \textbf{98.14} \\
Expert       & 91.23 & 87.03 &  \textbf{99.28} & 74.78 & 92.14 & 91.42 & 91.56 & 85.67 \\
LLM+Expert   & \textbf{91.92} & \textbf{88.23} &  97.14 & \textbf{79.31} & \textbf{94.28} & \textbf{95.60} & \textbf{93.92} & 92.69 \\
\bottomrule
\end{tabular}
\end{table}

\textbf{(5) Impact of expert knowledge.}
We further investigate the contribution of expert knowledge in guiding the diagnostic model, with results reported in Table~\ref{tab:knowledge_type}. This study compares model performance with and without the integration of expert-curated clinical knowledge. On AIBL, compared to LLM-based knowledge, using expert knowledge improves diagnostic AUC by 2.36\%; then, combining it with LLM and expert knowledge further improves the AUC by 1.20\%, demonstrating its added value over general-purpose LLM outputs. Similar improvements have been observed on the RENJI dataset. The strength of expert guidance lies in its ability to highlight clinically significant biomarkers that generic LLMs often overlook, as shown in Table~\ref{tab:knowledge} and ~\ref{tab:expert}.
Overall, expert knowledge offers clinical precision and contextual fidelity, particularly through fine-grained and task-specific criteria such as education-adjusted MMSE cutoffs (as seen in Table~\ref{tab:knowledge}). In contrast, the LLM contributes broader contextualization and comprehensive recall, linking findings to a wider range of medical knowledge, including rare phenotypes and underrepresented cases. That is, the expert helps anchor the reasoning process and correct potential hallucinations, while the LLM helps fill in gaps and broaden perspective. Together, they form a more balanced and complete diagnostic reasoning system.

\textbf{(6) Impact of loss weighting coefficients. }
To examine the influence of loss weighting coefficients on model performance, we conduct an ablation study on the AIBL test set by varying the coefficients $\lambda_{al}$ and $\lambda_{res}$ associated with the alignment loss ($\mathcal{L}_{al}$) and the restoration loss ($\mathcal{L}_{res}$). Similar to previous work on multi-loss optimization [1], we design five representative configurations to explore the trade-off between cross-modal alignment and modality-specific discriminability. As reported in Table~\ref{tab:lambda_ablation}, the balanced configuration ($\lambda_{al}=1.0$, $\lambda_{res}=1.0$) achieves the highest performance (91.92\% ACC, 88.23\% AUC), demonstrating the effectiveness of jointly optimizing both loss terms.

We observe that over-emphasizing alignment (Case B: $\lambda_{res}=0.1$) leads to a modest sensitivity gain (+0.79\%) but at the cost of a 3.89\% drop in specificity, indicating excessive feature homogenization. Conversely, prioritizing restoration (Case C: $\lambda_{al}=0.1$) slightly improves AUC (87.03\%) but degrades sensitivity, suggesting incomplete modality alignment. Removing either loss (Cases D and E) results in pronounced performance degradation, with the alignment-free setting (Case E) performing worst (89.23\% ACC). These findings highlight the interdependence between alignment and restoration objectives. A balanced weighting scheme mitigates the risks of over-smoothing or under-alignment, supporting more stable and generalizable multimodal learning.

\begin{table}[t]
\centering
\caption{Impact of loss term weights $\lambda_{al}$ and $\lambda_{res}$ on model performance (\%), evaluated on the AIBL test set.}
\label{tab:lambda_ablation}
\begin{tabular}{ccccccc}
\toprule
\textbf{Case} & \textbf{$\lambda_{al}$} & \textbf{$\lambda_{res}$} & \textbf{ACC} & \textbf{AUC} & \textbf{SEN} & \textbf{SPE} \\
\midrule
A & 1.0 & 1.0 & \textbf{91.92} & \textbf{88.23} &  97.14 & 79.31 \\
B & 1.0 & 0.1 & 91.28 & 86.67 & \textbf{97.93} & 75.42 \\
C & 0.1 & 1.0 & 90.76 & 87.03 & 96.28 & 74.77 \\
D & 1.0 & 0.0 & 91.09 & 84.80 &  95.83 & \textbf{81.76} \\
E & 0.0 & 1.0 & 89.23 & 86.28 & 93.65 & 78.91 \\
\bottomrule
\end{tabular}

\vspace{0.2cm}
\parbox{0.9\linewidth}{\footnotesize\textit{
Note: Optimal performance achieved with balanced weights (Case A). 
Excessive alignment focus (Case B) improves sensitivity at cost of specificity. 
Representation-only loss (Case E) shows weakest cross-modal alignment.}}
\end{table}

\begin{figure*}[t]
        \centering
        \includegraphics[width=\linewidth]{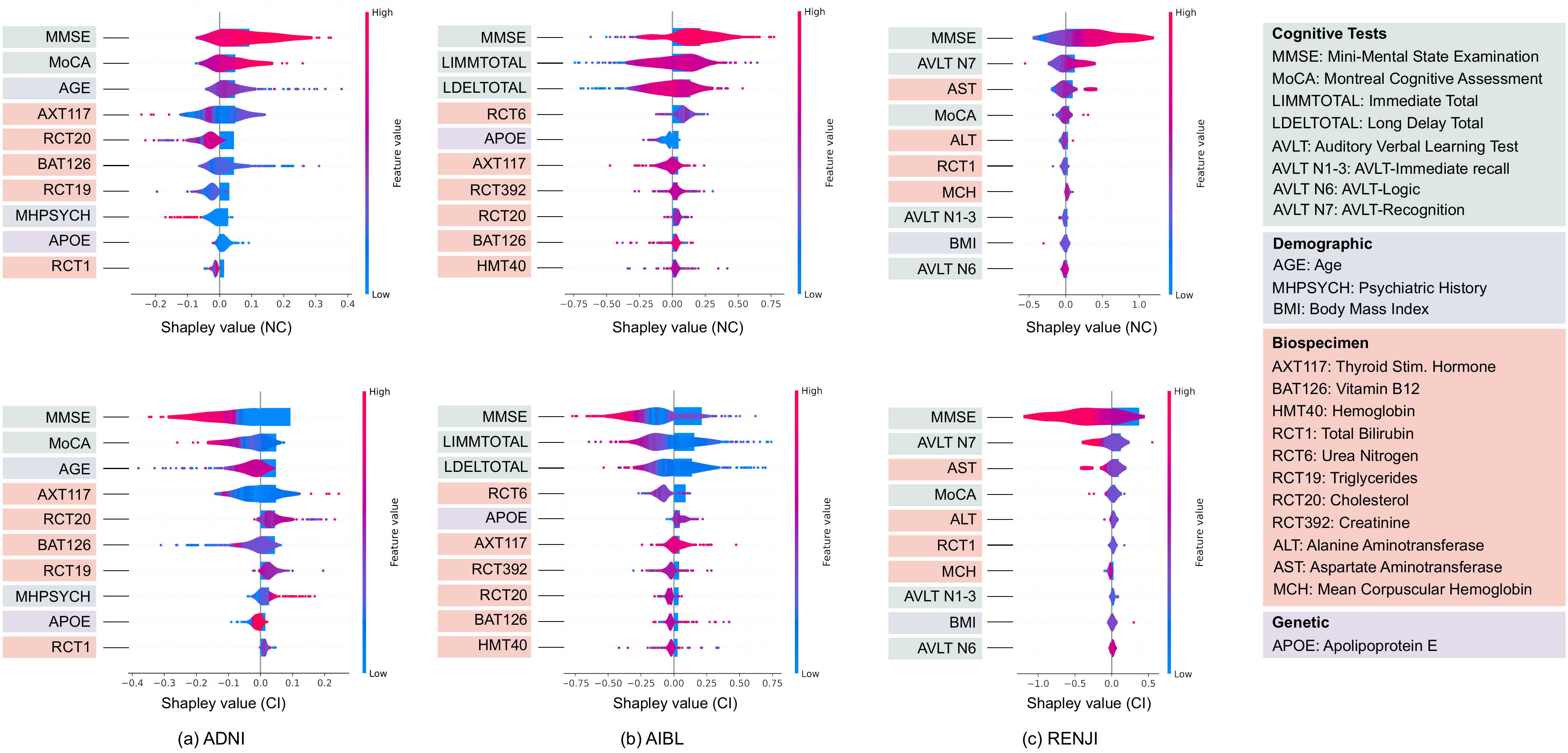}
	\caption{Shapley analysis is performed on cases from the ADNI, AIBL, and RENJI test sets. The results highlight the top ten numerical features contributing to the model's positive predictions for NC and CI labels, ranked by their mean Shapley values. These values represent the average contribution of each feature to the model's decisions, with features ordered from the highest to the lowest impact based on their influence.} 
\label{fig:shap}
\end{figure*}

\begin{figure*}[t]
\centering
\includegraphics[width=\linewidth]{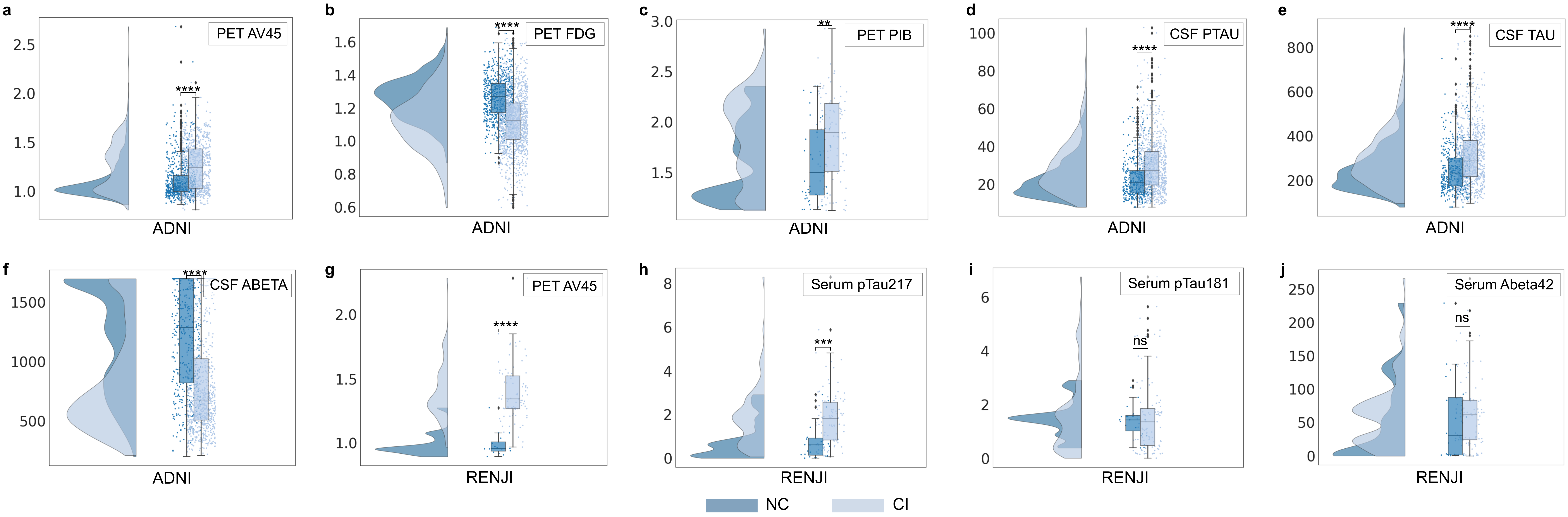}
\caption{Raincloud plots illustrate the distribution of biomarker values for model-predicted NC (dark blue) and CI (light blue). Model-predicted NC and CI groups were analyzed using t-tests on the ADNI dataset, focusing on (a) PET AV45 ($n=1406,t=−12.04,p=7.70 \times 10^{-32}$), (b) PET FDG ($n=1929,t=−18.43,p=5.19 \times 10^{-70}$), (c) PET PIB ($n=142,t=−3.17,p=1.85 \times 10^{-3}$), (d) CSF PTAU ($n=1257,t=−15.36,p=6.38 \times 10^{-49}$), (e) CSF TAU ($n=1257,t=8.11,p=1.16 \times 10^{-15}$), (f) CSF ABETA ($n=1257,t=8.66,p=1.48 \times 10^{-17}$) values, and on the RENJI dataset, focusing on (g) PET AV45 ($n=98,t=-7.22,p=1.23 \times 10^{-10}$), (h) Serum pTau217 ($n=124,t=-3.65,p=3.87 \times 10^{-4}$), (i) Serum pTau181 ($n=124,t=-0.33,p=7.44 \times 10^{-1}$), (j) Serum ABETA42 ($n=124,t=-0.79,p=4.33 \times 10^{-1}$) values.} 
\label{fig:biomarker}
\end{figure*}

\subsubsection{Factor Analysis}

Shapley analysis~\cite{NIPS2017_7062} is performed on the ADNI, AIBL, and RENJI test sets to identify the numerical features that most significantly influenced the model's diagnostic decisions (Fig.~\ref{fig:shap}). Across all three datasets, the MMSE score consistently ranks among the most influential features. For both ADNI and AIBL, MMSE, Thyroid Stimulating Hormone, Vitamin B12 levels, and the presence of APOE4 alleles are selected consistently among the top ten factors. These findings align with clinical studies that emphasize the strong association of MMSE scores and other key biomarkers with cognitive impairment and AD diagnosis.

In the RENJI dataset, Auditory Verbal Learning Test (AVLT) scores and their subscores emerge in the top ten factors, closely aligned with the knowledge provided by physicians and clinical studies~\cite{moradi2017rey}. AVLT is a test designed to assess auditory memory, which is valuable for detecting early-stage cognitive decline. Notably, elevated serum aspartate aminotransferase (AST) levels, previously associated with AD in comparative studies~\cite{li2022increased}, and Body Mass Index (BMI), linked to cognitive impairment in other research~\cite{qu2020association, ren2021associations}, also featured prominently in the RENJI top ten features.

\begin{table*}[t]
\begin{center}
\caption{Knowledge generated by Different LLMs for sampled factors.
}
\label{tab:llm_knowledge}
\begin{tabular}{p{30pt}p{60pt}p{380pt}}
\toprule
\textbf{Factor} & \textbf{LLM} & \textbf{Knowledge} \\
\cmidrule(r){1-3}
MMSE & GPT4o~\cite{achiam2023gpt} & A lower MMSE score indicates poorer cognitive function, which is often seen in individuals with Alzheimer's disease. \\
\cmidrule(r){2-3}
& Llama3~\cite{grattafiori2024llama} &  Lower MMSE scores are associated with poorer cognitive function and an increased risk of AD. Low MMSE scores may indicate a higher risk of AD.  \\
\cmidrule(r){2-3}
& Med-llama3~\cite{xie2024me} &  The Mini-Mental State Examination (MMSE) is a brief cognitive test that is widely used in clinical practice. Its diagnostic accuracy for Alzheimer's disease (AD) in the memory clinic population is unclear. We aimed to determine the diagnostic accuracy of the MMSE in a memory clinic population. The study was a secondary analysis of the Australian Imaging, Biomarkers and Lifestyle (AIBL) Study of Ageing.  \\
\cmidrule(r){2-3}
& HuatuoGPT-V~\cite{chen2024huatuogpt} & A lower MMSE score indicates poorer cognitive performance, which may be associated with an increased likelihood of AD diagnosis. However, the relationship between MMSE scores and AD diagnosis can also depend on other factors such as age, education level, and severity of symptoms.  \\
\cmidrule(r){1-3}
APOE & GPT4o~\cite{achiam2023gpt} & The presence of the APOE-$\epsilon$4 allele increases an individual's genetic risk for developing Alzheimer's disease (AD), influencing both the likelihood and the potential age of onset. \\
\cmidrule(r){2-3}
& Llama3~\cite{grattafiori2024llama} & Carrying the APOE-$\epsilon$4 allele is a significant genetic risk factor for Alzheimer's disease (AD); individuals with one or two copies of the APOE-e4 gene have an increased likelihood of developing AD. \\
\cmidrule(r){2-3}
& Med-llama3~\cite{xie2024me} & The apolipoprotein E (APOE) gene is a major risk factor for Alzheimer's disease (AD). The APOE $\epsilon$4 allele is associated with a 3- to 4-fold increased risk of developing AD, while the APOE $\epsilon$2 allele is associated with a reduced risk. The aim of this study was to systematically review the evidence on the relationship between APOE and AD diagnosis, and to perform a meta-analysis of the diagnostic accuracy of APOE genotyping in AD. \\
\cmidrule(r){2-3}
& HuatuoGPT-V~\cite{chen2024huatuogpt} & The APOE gene has three main variants: $\epsilon$2, $\epsilon$3 (most common), and $\epsilon$4. The presence of the $\epsilon$4 allele is associated with an increased risk for Alzheimer's disease (AD) diagnosis; individuals homozygous or heterozygous for the $\epsilon$4 variant have a higher likelihood of developing AD compared to those without this genetic factor. \\
\bottomrule
\end{tabular}
\end{center}
\end{table*}

\subsubsection{Biomarker-Level Validation}

To enhance clinical accessibility, our model relies on readily available non-pathological factors and validates its effectiveness through biomarker-level analysis. As shown in the raincloud plots (Fig.~\ref{fig:biomarker}), we visualize the distributions of biomarker values for individuals predicted as cognitively normal (NC, dark blue) and cognitively impaired (CI, light blue). Each raincloud plot comprises three elements:
(i) the ``cloud", representing a smoothed density curve of the value distribution;
(ii) the ``rain", showing individual data points jittered horizontally to reflect raw cohort-level observations;
(iii) the boxplot, summarizing the median, quartiles, and outliers.
This integrated visualization reveals strong concordance between model predictions and known neuropathological markers. In the ADNI dataset, predictions correlate with PET biomarkers (AV45, FDG, PIB) and CSF biomarkers (PTAU, TAU, Abeta), while in the RENJI cohort, they align with PET AV45 and serum pTau217 levels. Notably, both predicted classes and biomarker distributions show clear separation between NC and CI groups (all inter-group comparisons: $p \leq 0.001$), confirming the biological plausibility of the model’s outputs. Among these, serum pTau217 exhibits superior discriminative power compared to pTau181 and Abeta42, consistent with its established utility in detecting cerebral amyloidosis and predicting cognitive decline~\cite{lehmann2024clinical}.

\subsubsection{Comparison of Different LLMs for Knowledge Generation}
We conduct a comprehensive evaluation of several state-of-the-art LLMs, including GPT-4o~\cite{achiam2023gpt}, HuatuoGPT-Vision~\cite{chen2024huatuogpt}, Llama3~\cite{grattafiori2024llama}, and Med-Llama3~\cite{xie2024me}.
Qualitative examples of knowledge generation for representative factors (MMSE and APOE) are detailed in Table~\ref{tab:llm_knowledge}. HuatuoGPT-V consistently provides the most clinically relevant and comprehensive descriptions of both MMSE and APOE factors, effectively incorporating essential clinical context, individual variability, and explicit differentiation among genetic variants. GPT-4o, although succinct and accurate, lacks certain contextual nuances and variant-specific details essential for clinical applications. Meanwhile, Med-Llama3 and Llama3 exhibit notable limitations, including overly academic phrasing and insufficient detail, which significantly reduces their practical utility. 

We further employ automated metrics provided by the third-party LLM Deepseek~\cite{guo2025deepseek} to evaluate the quality of the generated knowledge, supplemented by expert spot-check validation. This assessment focuses on their capability to generate clinically accurate descriptions of disease-related factors across four critical dimensions: terminology accuracy, clinical relevance, practical utility, and consistency. Quantitative comparisons of the models’ knowledge element outputs are systematically summarized in Table~\ref{tab:model-evaluation}. Although GPT-4o achieved the highest quantitative performance, its computational demands made it unsuitable for integration into our lightweight diagnostic framework. In contrast, HuatuoGPT-V demonstrated superior clinical relevance and consistency compared to Llama3 and Med-Llama3 (15$\sim$69\% improvement). Furthermore, additional experiments on the AIBL dataset (Table~\ref{tab:ablation_llm}) confirmed minimal impact on diagnostic performance ($\leq 1$\% AUC difference and the same accuracy) between GPT-4o and HuatuoGPT-V, validating our selection of HuatuoGPT-V for its balanced performance and practical deployability. 

These combined quantitative results and qualitative assessments underscore the superiority of HuatuoGPT-V, substantiating its optimal balance of clinical precision, explanatory depth, and deployment efficiency.

\setlength{\tabcolsep}{3pt}
\begin{table}[t]
\centering
\scriptsize
\caption{Precision scoring (\%) of Alzheimer's Disease descriptions generated by different LLMs and evaluated by Deepseek~\cite{guo2025deepseek}.}
\label{tab:model-evaluation}
\begin{tabular}{lccccc}
\toprule
\multirow{2}{*}{\textbf{Model}} & 
\multicolumn{4}{c}{\textbf{Evaluation Dimensions}} & \multirow{2}{*}{\textbf{Total}} \\
\cmidrule(r){2-5}
& \textbf{Terminology} & \textbf{Relevance} & \textbf{Utility} & \textbf{Consistency} & \\
\midrule
GPT-4o~\cite{openai2023gpt4} & \textbf{98.6} & \textbf{99.8} & \textbf{96.2} & \textbf{99.1} & \textbf{98.5} \\
Llama3~\cite{grattafiori2024llama} & 83.2 & 71.4 & 63.8 & 78.3 & 73.8 \\
Med-Llama3~\cite{xie2024me} & 27.5 & 18.3 & 12.6 & 23.7 & 20.0 \\
HuatuoGPT-V~\cite{chen2024huatuogpt} & 91.7 & 93.5 & 84.9 & 87.4 & 89.7 \\
\bottomrule
\end{tabular}

\vspace{3mm}
\footnotesize
\begin{tabular}{p{0.95\linewidth}}
\textsuperscript{1}Weighting: Terminology (35\%), Relevance (30\%), Utility (25\%), Consistency (10\%) \\
\textsuperscript{2}Total score = $\sum$(Dimension score $\times$ Dimension weight)
\end{tabular}
\end{table}

\begin{table}[t]
\begin{center}
    \caption{Ablation results (\%) on the AIBL test dataset using different LLM-based knowledge descriptions. Med-Llama3 is excluded due to poor performance in Table~\ref{tab:model-evaluation} (Best in bold).
    }
    \label{tab:ablation_llm}
    \begin{tabular}{lllll}
    \toprule
        \textbf{Task} & \textbf{ACC} & \textbf{AUC} & \textbf{SEN} & \textbf{SPE} \\
    \cmidrule(r){1-5}
        Llama3~\cite{grattafiori2024llama} & 86.87 & 81.11 &  94.37 & 67.86 \\
        HuatuoGPT-V~\cite{chen2024huatuogpt} & \textbf{91.92} & 88.23 &  \textbf{97.14} & 79.31 \\
        GPT-4o~\cite{openai2023gpt4} & \textbf{91.92} & \textbf{89.24} & 95.71 & \textbf{82.76}  \\
    \bottomrule
    \end{tabular}
  \end{center}
\end{table}

\begin{table}[t]
\begin{center}
    \caption{Comparison of diagnostic accuracy (\%) on the AIBL test dataset across text-only, vision-language, and our specialized multimodal methods. (Best in bold).
    }
    \label{tab:ablation_llm_holodx}
    \begin{tabular}{lllll}
    \toprule
        \textbf{Task} & \textbf{ACC} & \textbf{AUC} & \textbf{SEN} & \textbf{SPE} \\
    \cmidrule(r){1-5}
        LLM (text-only) & 75.76 & 60.64 &  97.14 & 24.14  \\
        VLLM (text + 2D slice) & 76.77 & 60.01 &  \textbf{98.60} & 21.43   \\
        HoloDx (ours) & \textbf{91.92} & \textbf{88.23} &  97.14 & \textbf{79.31}  \\
    \bottomrule
    \end{tabular}
  \end{center}
\end{table}

\begin{table}[t]
\setlength{\tabcolsep}{3pt}
\centering
\caption{Comparison of model efficiency and complexity on the AIBL dataset.}
\label{tab:efficiency_full}
\begin{tabular}{@{}lrrrr@{}}
\toprule
\multirow{2}{*}{\textbf{Method}} & \textbf{Training Time} & \textbf{GPU Mem} & \textbf{Parameters} & \textbf{Inference Time} \\ 
 & \textbf{/Epoch (min)} & \textbf{(GB)} & \textbf{(MB)} & \textbf{/Sample (ms)} \\ 
\midrule
AD-Trans~\cite{yu2024transformer} & 2.6 & 4.7 & 201.5  & 135.2 \\
Alifuse~\cite{chen2024alifuse}    & 7.4 & 14.1 & 608.7  & 178.1 \\
Backbone \tiny{(our BB)} & 6.4 & 9.7 & 228.2  & 223.6 \\
BB+KL & 8.7\tiny{~(+35.9\%)} & 12.1\tiny{~(+24.7\%)} & 276.0\tiny{~(+20.9\%)} & 231.7\tiny{~(+3.6\%)} \\
BB+KL+MEM   & 10.5\tiny{~(+64.1\%)} & 13.1\tiny{~(+35.0\%)} & 303.8\tiny{~(+33.1\%)} & 232.9\tiny{~(+4.2\%)} \\
\bottomrule
\end{tabular}
\end{table}

\subsubsection{Direct LLM/VLLM Diagnosis vs. Specialized Multimodal Modeling}
Our HoloDx is designed to address key limitations of generic LLM or vision-augmented (VLLM) methods by deeply integrating clinical expertise and imaging information. Specifically, the framework incorporates 3D brain images to capture structural and metabolic abnormalities, embeds domain-specific clinical data and knowledge aligned with AD pathology, and leverages LLMs as adaptive knowledge retrievers, rather than standalone diagnostic agents. This design enables our model to deliver clinically grounded, individualized interpretations while ensuring efficiency and generalization.

To validate the superiority of our specialized method, we conduct a comparative evaluation reported in Table~\ref{tab:ablation_llm_holodx}. Pure text-based LLM diagnosis, which relies solely on symptom descriptions without imaging data, exhibits 16.16\% lower accuracy compared to HoloDx. The VLLM achieves moderate performance (76.77\% accuracy on AIBL test dataset) but remains constrained by its dependence on 2D visual features and general medical knowledge, lacking the capacity to model 3D neuroimaging patterns or domain-specific clinical context. In contrast, our multimodal design achieves 91.92\% diagnostic accuracy, demonstrating a statistically significant improvement over generic LLM/VLLM methods.

\subsubsection{Comparison of Model Efficiency and Complexity}
To address concerns regarding computational and storage demands, we conducted comprehensive efficiency measurements across all methods using a standardized hardware platform under identical software conditions. As systematically quantified in Table~\ref{tab:efficiency_full}, our complete framework incorporating both knowledge and memory injection modules exhibits a measured increase in resource requirements while maintaining clinically viable performance characteristics. The knowledge injection component introduces moderate overhead—35.9\% longer training times and 24.7\% higher memory consumption—primarily attributable to its cross-modal attention mechanisms and the computational cost of knowledge alignment during optimization. When augmenting this foundation with our memory injection module, we observe further increases in training time (64.1\% total increase over our backbone) and parameter count (33.1\% expansion), though crucially with only marginal impact on inference latency (4.2\% increase).
While the training time per epoch increases to 10.5 minutes and peak memory utilization reaches 13.1GB, these remain within practical boundaries for hospital computing infrastructure. The measured resource increases are clinically justified by our method's 5.05\% AUC improvement over baseline (Table~\ref{tab:ablation}(b).
When compared to contemporary approaches, our solution demonstrates superior memory efficiency relative to Alifuse \cite{chen2024alifuse} while maintaining competitive training durations.

\section{Conclusion and Discussion}
In this paper, we propose a knowledge- and data-driven method for AD diagnosis, integrating expert-provided and LLM-generated knowledge with multimodal imaging and textual data. At the core of our framework are two key modules: (1) the knowledge injection module, which incorporates domain-specific knowledge from experts and LLM-generated insights, enriching the model’s understanding of AD; and (2) the memory injection module, which stores and retrieves relevant past patient information, enabling the model to leverage historical knowledge for more informed decision-making. To further enhance model performance, we employ contrastive and restorative learning strategies to effectively align and fuse multimodal data with knowledge.

Our experiments on multiple AD datasets show that the proposed method consistently outperforms recent baselines, achieving superior classification performance. Also, we visualize the final diagnosis through Shapley analysis and raincloud plots. Shapley values allow us to identify the most important non-imaging contributions, shedding light on how cognitive and genetic factors influence the model's predictions. Raincloud plots further demonstrate the alignment between model-predicted diagnoses and clinical biomarkers, ensuring that the model's decisions are consistent with established diagnostic criteria. In future work, we plan to improve the generalization and cross-domain applicability of our model, ensuring it performs effectively across different datasets. Additionally, we will focus on expanding and updating the knowledge base, integrating more expert-provided and LLM-generated information to enhance the model's diagnostic capabilities.

\bibliographystyle{IEEEtran}
\bibliography{refs}

\end{document}